\documentclass {article} 
\usepackage{arxiv}
\usepackage{booktabs} 
\usepackage{algorithm}%
\usepackage{algpseudocode}%
\usepackage{hyperref}
\usepackage{amsmath}
\usepackage{xcolor}
\usepackage{xspace}
\usepackage{soul}
\newcommand{\ignore}[1]{}
\usepackage[textsize=tiny]{todonotes}
\usepackage{graphicx}
\usepackage{boldline}
\usepackage[export]{adjustbox}
\usepackage[caption=false]{subfig}
\usepackage[font=small]{caption}
\usepackage{adjustbox}
\usepackage{enumerate}
\usepackage{amsfonts}
\usepackage{colortbl}
\usepackage{float}
\usepackage{rotating}
\usepackage{setspace}
\usepackage[official]{eurosym} 
\usepackage[flushleft]{threeparttable}
\usepackage{enumitem}
\usepackage{array}
\title{Hybrid Neuro-Evolutionary Method for Predicting Wind Turbine Power Output}

\author{
 Mehdi Neshat \\
  Optimization and Logistics Group\\
  School of Computer Science\\
  The University of Adelaide\\
   Australia \\
  \texttt{mehdi.neshat@adelaide.edu.au} \\
   \And
   Meysam Majidi Nezhad\\
Department of Astronautics\\
Electrical and Energy Engineering (DIAEE)\\
	Sapienza University of Rome\\
	 Italy\\
	 \texttt{meysam.majidinezhad@uniroma1.it} \\
   \And
Ehsan Abbasnejad \\
  The Australian Institute for Machine Learning\\
   The University of Adelaide\\
   Australia \\
  \texttt{ehsan.abbasnejad@adelaide.edu.au} \\
  \And
  Daniele Groppi \\
  Department of Astronautics\\
Electrical and Energy Engineering (DIAEE)\\
	Sapienza University of Rome\\
	 Italy\\
  \texttt{daniele.groppi@uniroma1.it} \\
  \And
  Azim Heydari \\
  Department of Astronautics\\
Electrical and Energy Engineering (DIAEE)\\
	Sapienza University of Rome\\
	 Italy\\
  \texttt{ azim.heydari@uniroma1.it} \\
  \And
  Lina Bertling Tjernberg \\
  School of Electrical Engineering and Computer Science\\
  KTH Royal Institute of Technology Stockholm\\
  Sweden\\
    \texttt{Linab@kth.se} \\
  \And
  Davide Astiaso Garcia \\
  Department of Planning, Design\\
  and Technology of Architecture\\
  Sapienza University of Rome\\
  Italy\\ 
    \texttt{davide.astiasogarcia@uniroma1.it} \\
  \And
  Bradley Alexander \\
  Optimization and Logistics Group\\
  School of Computer Science\\
  The University of Adelaide\\
   Australia \\
  \texttt{bradley.alexander@adelaide.edu.au} \\
  \And
  Markus Wagner \\
  Optimization and Logistics Group\\
  School of Computer Science\\
  The University of Adelaide\\
   Australia \\
  \texttt{markus.wagner@adelaide.edu.au} 
  }

\begin{document}

\maketitle
\doublespacing
\begin{abstract}
Reliable wind turbine power prediction is imperative to the planning, scheduling and control of wind energy farms for stable power production. In recent years Machine Learning (ML) methods have been successfully applied in a wide range of domains, including renewable energy. However, due to the challenging nature of power prediction in wind farms, current models are far short of the accuracy required by industry. In this paper, we deploy a composite ML approach--namely a hybrid neuro-evolutionary algorithm--for accurate forecasting of the power output in wind-turbine farms. We use historical data in the supervisory control and data acquisition (SCADA) systems as input to estimate the power output from an onshore wind farm in Sweden. At the beginning stage, the k-means clustering method and an Autoencoder are employed, respectively, to detect and filter noise in the SCADA measurements. Next, with the prior knowledge that the underlying wind patterns are highly non-linear and diverse, we combine a self-adaptive differential evolution (SaDE) algorithm as a hyper-parameter optimizer, and a recurrent neural network (RNN) called Long Short-term memory (LSTM) to model the power curve of a wind turbine in a farm. Two short time forecasting horizons, including ten-minutes ahead and one-hour ahead, are considered in our experiments. We show that our approach outperforms its counterparts. 

\end{abstract}


\sloppy

\section{Introduction}\label{sec:Introduction}

 Renewable wind energy is an established but fast-growing technology for the sustainable production of energy at scale. With falling costs and large-scale production of generators the deployment of wind energy is accelerating, for example, gross installations in the EU of the onshore and offshore wind farm were 0.3 GW in 2008 and increased to 3.2 GW in 2017~\cite{komusanac2019wind}. With such large increases in the deployment of wind energy forecasting of the power output of installed wind turbines is becoming vitally important. 
However, because local wind environments in wind farms are complex, and because the responses of wind turbines are non-linear and dependent on the condition of the turbine, wind power forecasting is a challenging problem~\cite{lydia2014comprehensive}.

Wind power forecasting is fundamental to the effective integration of wind farms into the power grid. For a single turbine, the following equation describes power output: $P$. 
\begin{equation}\label{eq-curve Power}
P=\frac{1}{2}\rho\pi R^2C_p u^3
\end{equation}
The terms $\rho$, $R$ and $u$ are the density of air, the rotor radius and the wind speed respectively. Meanwhile, $C_p$ is the power coefficient the proportion of available power the turbine is able to extract. The theoretical estimation of wind turbine power is depicted by Equation \ref{eq-curve Power}.
This equation describes a smooth s-shaped power curve that resembles a logistic function with wind on the x-axis. 
However, because of the variable nature of wind and complex dynamics within and between turbines, the real power output of individual wind turbines is not precisely described by this curve~\cite{saleh2016hybrid,morshedizadeh2017improved}. An alternative model for each wind turbine in a farm can be derived by fitting observations to field data to derive more realistic models~\cite{lydia2014comprehensive}. On top of the task of modelling wind turbine power in response to {\em{current}} wind conditions managers of wind farms also need to forecast {\em{future}} power output based on current conditions. Recent work has used complex data-driven models such as artificial neural networks (ANNs) to forecast turbine output with some degree of  accuracy~\cite{lin2020wind,rodriguez2020very,nielson2020using}. 
 
In this paper, we propose an integrated approach that couples self-adaptive differential evolution with ANNs for accurate short term wind power forecasting. The input features used in our modelling are current wind, speed, current wind direction and (in some models) current power output. 

\begin{figure*}[tbp]
 \centering
  \includegraphics[width=0.8\textwidth]{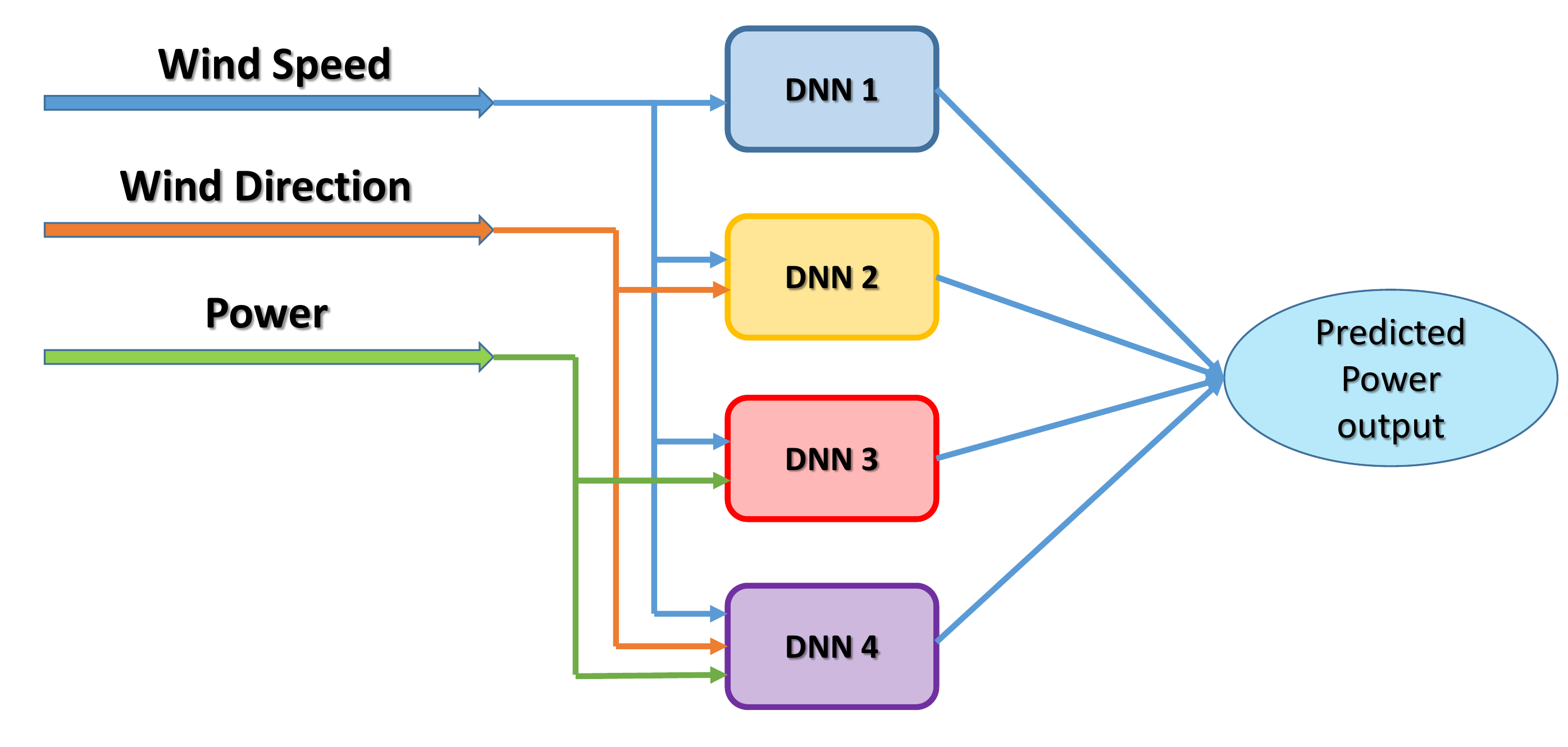}
   \caption{The proposed four different independent forecasting models.The applied power as an input is the current generated power by the wind turbine.  }
   \label{fig:all_stsrems}
 
  \end{figure*}

The main contributions of this paper are:
\begin{enumerate}
\item a new hybrid Neuro-evolutionary method (SaDE-LSTM) for short term wind turbine power output forecasting that combines self-adaptive differential evolution (SaDE)~\cite{qin2005self} to act on a recurrent deep neural network~\cite{hochreiter1997long} with two forecasting horizons of ten-minutes and one-hour; 

  \item an implementation of an advanced data filtering technique is implemented on the training observations 
  (from SCADA data) using K-means clustering~\cite{macqueen1967some} 
  and autoencoder neural-networks~\cite{vincent2010stacked} to detect outliers;

  \item a comparison of the performance of the models with raw and clean SCADA datasets for assessing the impact of the outlier detection method;
  
  \item a comparison of the performance or four 
  forecasting models trained to act on different subsets of SCADA inputs. These sets are wind speed for model one; wind speed and wind direction for model two; wind speed and current power output for model three; and, finally, wind speed, wind direction and current power output for model four. 
  Figure \ref{fig:all_stsrems} illustrates the models and their inputs; 
  
  \item finally, as there is not a straightforward theory with regard to the design and tune the hyper-parameters of an LSTM network~\cite{hu2018nonlinear}, we tune the model structure and hyper-parameters using: grid search;  the Gray Wolf Optimizer~\cite{mirjalili2014grey} (GWO) method; Differential Evolution~\cite{storn1997differential} (DE) algorithm; and covariance matrix adaptation evolution strategy~\cite{hansen2004evaluating} (CMA-ES). 
\end{enumerate}

The remainder of this article is organized as follows: Section~\ref{sec:Related-works} reviews current approaches to building wind power forecasting models. Section~\ref{sec:SCADA-data} describes the features of the SCADA datasets employed in this research, collected from 42-months of high-frequency monitoring of onshore wind turbines. Section~\ref{sec:preprocessing} describes the new
outlier detection method used in this work. 
 Section~\ref{sec:Methodology} describes the experimental methodology. 
After that, the power prediction results trained by datasets of both raw and clean SCADA datasets are demonstrated. Finally, Section \ref{sec:Conclusions} summarises 
the contributions of this work. 

\section{Related Work}
\label{sec:Related-works}

Some work in this field combined physical models with numerical weather prediction (NWP) to forecast wind turbine output~\cite{lange2006physical}. While such methods are fast, they also exhibit low accuracy.

More accurate predictions can be obtained by using statistical methods to model the relationship between the inputs of the system and the corresponding outputs. Commonly applied statistical methods include time-series methods~\cite{bergmeir2012use,kavasseri2009day}, machine learning  methods~\cite{bhaskar2012awnn,he2018probability}, the  persistence method~\cite{brown1984time} and Kalman filtering~\cite{zuluaga2015short}.

One set of well-known time series forecasting methods in the wind power prediction is autoregressive modelling using: autoregressive moving average (ARMA), autoregressive integrated moving average (ARIMA), and a fractional version of ARIMA (f-ARIMA)~\cite{yuan2017wind}. Both ARMA and ARIMA models can capture short-range correlations between inputs and outputs, and the f-ARIMA method is well adapted for representing the time series data with long memory characteristics~\cite{kavasseri2009day}. 

Another effective technique for forecasting time series data is recurrent neural networks (RNNs). Olaofe et al.~\cite{olaofe2012wind} applied  RNNs to predict the wind turbine power output one-day ahead.
However, the applied 'tanh' activation function used in this work lead to disappearing and exploding gradients, which lead to difficultly in training an accurate model~\cite{bengio1994learning}. Long short-term memory networks (LSTMs) were introduced in~\cite{hochreiter1997long}, partly, to help avoid these issues. LSTMs can learn the correlations carried in time series data with some accuracy.  In~\cite{Zhu2017LSTM}, LSTMs were employed for short term predictions of wind power. That study showed that LSTMs could outperform traditional ANNs and support vector machines (SVMs) in terms of prediction accuracy. 
A combination of principal component analysis (PCA) and an LSTM forecasting model was proposed in~\cite{xiaoyun2016short}, and compared with BP neural network and an SVM model. Their results showed that the PCA-LSTM framework results produced higher forecasting accuracy than other methods. 
Recently, Erick et al.~\cite{lopez2018wind} defined a new architecture for wind power forecasting composed of LSTM blocks replacing the hidden units in the Echo State Network (ESN). They also used quantile regression to produce a robust estimation of the proposed forecast target.  
Finally, u et al.~\cite{yu2019lstm}, used an LSTM with an enhanced forget-gate network model (LSTM-EFG) combined with a Spectral Clustering method to forecasting wind power with a considerable increase in accuracy. 

However, none of the above work used an automated method for tuning the hyperparameters of their ANN models. Such automatic tuning helps with porting the models to a new setting and makes it possible to compare modelling approaches more rigorously. 
 
Some recent works that have used hyper-parameter tuning include  Qin et al.~\cite{Qin2015hybrid} who used the Cuckoo Search Optimization (CSO) method to improve performance of a Back Propagation Neural Network (BPNN) by adjusting the connection weights. They reported that the accuracy of the proposed hybrid model improves on other methods for predicting the wind speed time series. Shi et al.~\cite{shi2017direct} used the dragonfly algorithm (DA) to tune RNN hyperparameters for wind power forecasting. 

In another recent work, Peng et al.~\cite{peng2018effective} used  Differential Evolution (DE)  to optimise  LSTM parameters, and the reported results indicated that the hybrid DE-LSTM model is able to outperform traditional forecasting models in terms of prediction accuracy. More recently, Neshat et al.~\cite{neshat2020evolutionary} forecast time series with online hyperparameter tuning of an LSTM  model using CMA-ES. This work improves on earlier work by systematically comparing the impact of tuning strategies, model input sets, and data pre-processing on prediction performance. These comparisons define some of the search landscape in the parameter space of these algorithms. 
 \begin{figure*}[tbp]
 \centering
  \includegraphics[width=\textwidth]{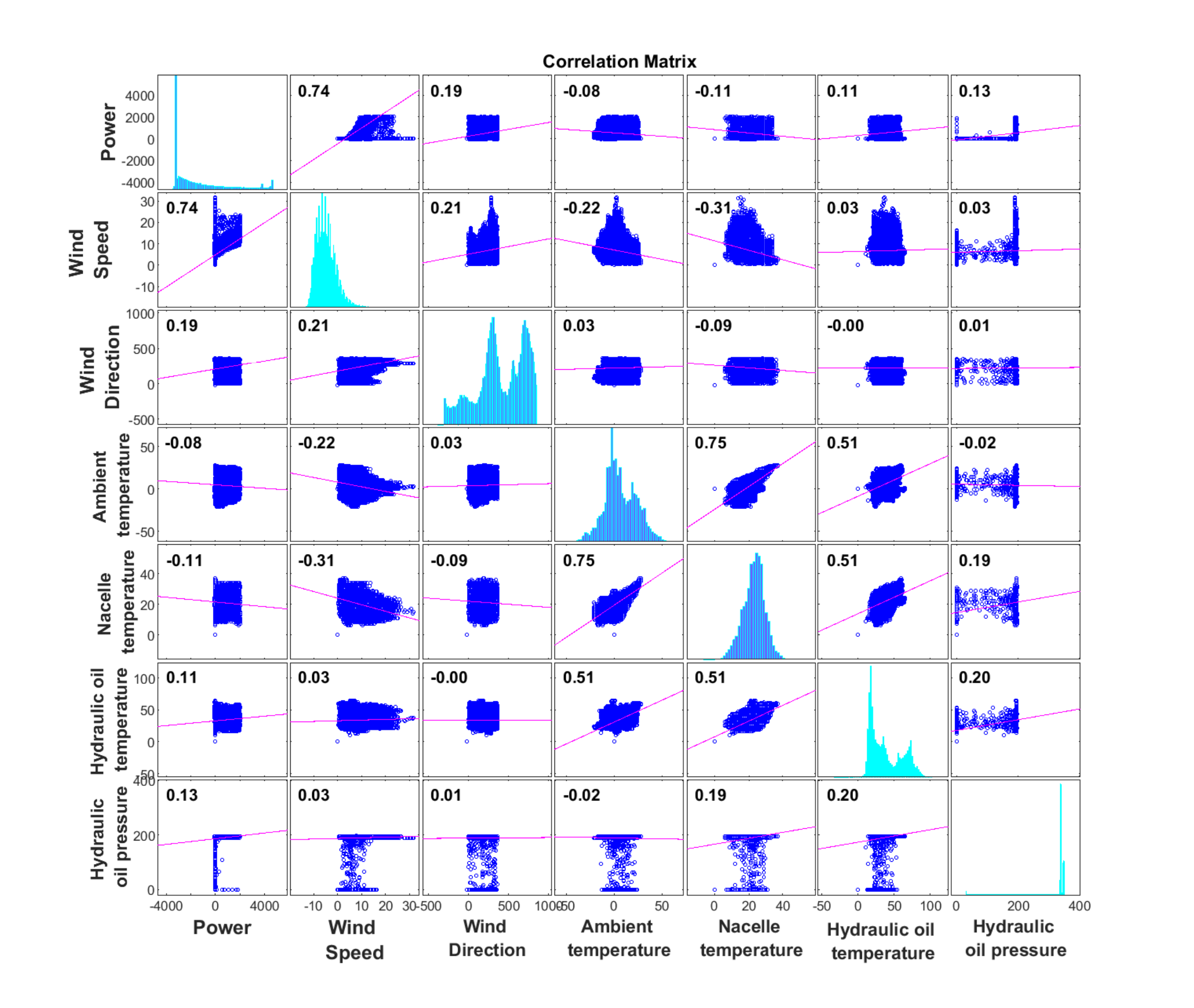}
   \caption{The Pearson's linear correlation coefficients between all pairs of the wind turbine data (SCADA). The correlation  plot shows that wind speed, wind direction and Power are highly correlated.   }
   \label{fig:correlation}
 
  \end{figure*}

\section{SCADA data description and analysis} \label{sec:SCADA-data}

The data applied in this research comes from six turbines of
one onshore wind farm in the north-western
Europe (Sweden)~\cite{huang2019wind}. For each turbine, 42 months of data are available from January of 2013 to June of 2016, which includes 10-minute interval operation data and a log file. In addition, faults and maintenance information are also stored. As a research sample, we select and investigate the SCADA data from the sixth turbine in the wind farm in this paper. For evaluating and analysing how power output correlates with other SCADA features, seven features are chosen including wind speed, wind direction, ambient temperature, Nacelle temperature, Hydraulic oil temperature and Hydraulic oil pressure. These are the most recommended SCADA features for power prediction from~\cite{lin2020wind}. Pearson's linear correlation coefficients between all pairs of the wind turbine data features can be seen in Figure~\ref{fig:correlation}. The highest correlation is between Power output and wind speed as well as wind direction. Therefore, we select the wind speed and wind direction as the inputs of the ANN with the network output being generated power. 
The diagonal in Figure~\ref{fig:correlation} shows the distributions of each variable, including power. These distributions show some outliers, which might pose challenges for modelling. It is also of note that there are some negative values for produced power; this is caused by stationary turbines spinning up. 

\begin{figure*}[tbp]
 \centering
  \includegraphics[width=0.7\textwidth]{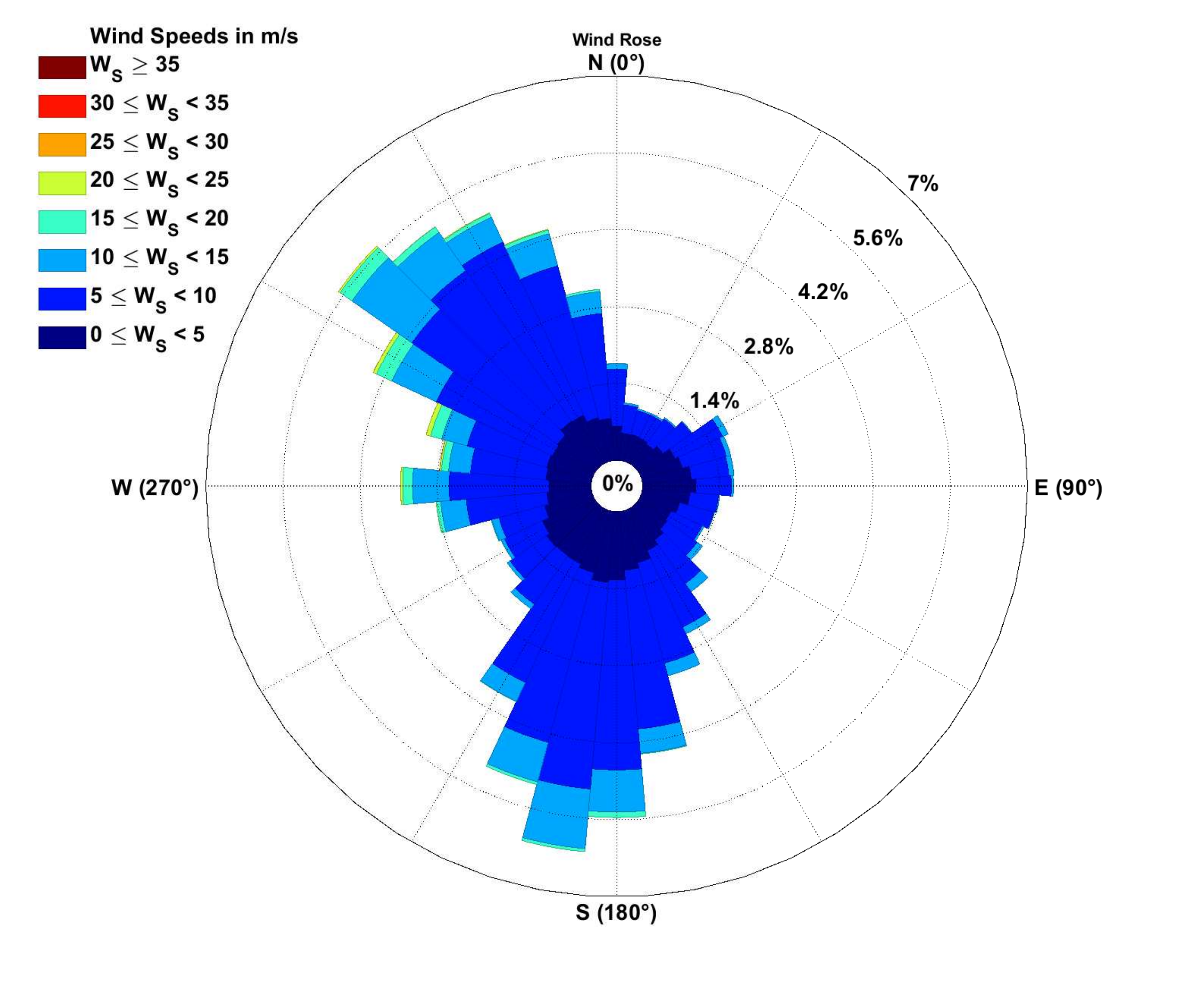}
   \caption{ a large view of how the wind speed and wind direction are distributed at the wind farm (Sweden) from 2013 to 2016 (June).  }
   \label{fig:wind_Rose}
 
  \end{figure*}

Figure \ref{fig:wind_Rose} 
is the wind rose for the wind far. It shows that the dominant wind direction is  North-west, and a secondary prevailing direction is South-east. However, there are also occasional West winds.
 \begin{figure*}[tbp]
  \subfloat[]{
  \includegraphics[width=0.95\textwidth]{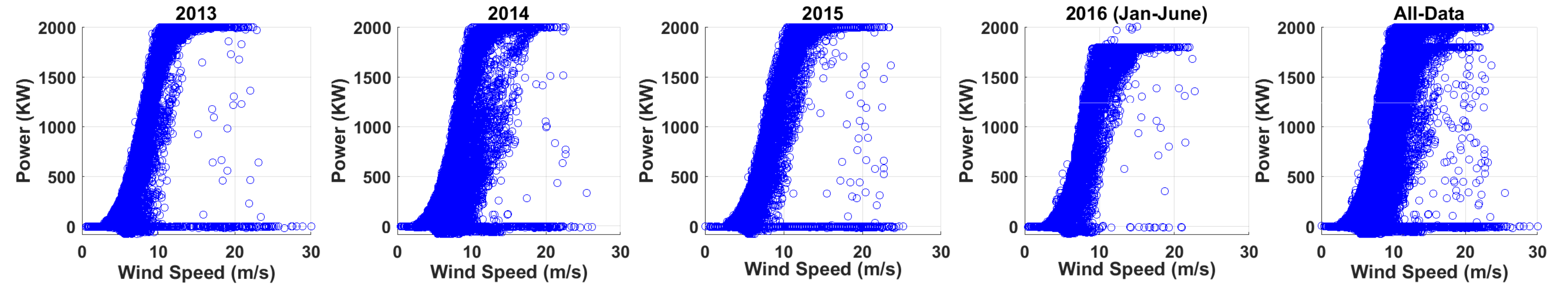}}\\
  \subfloat[]{
\includegraphics[clip,width=0.31\columnwidth]{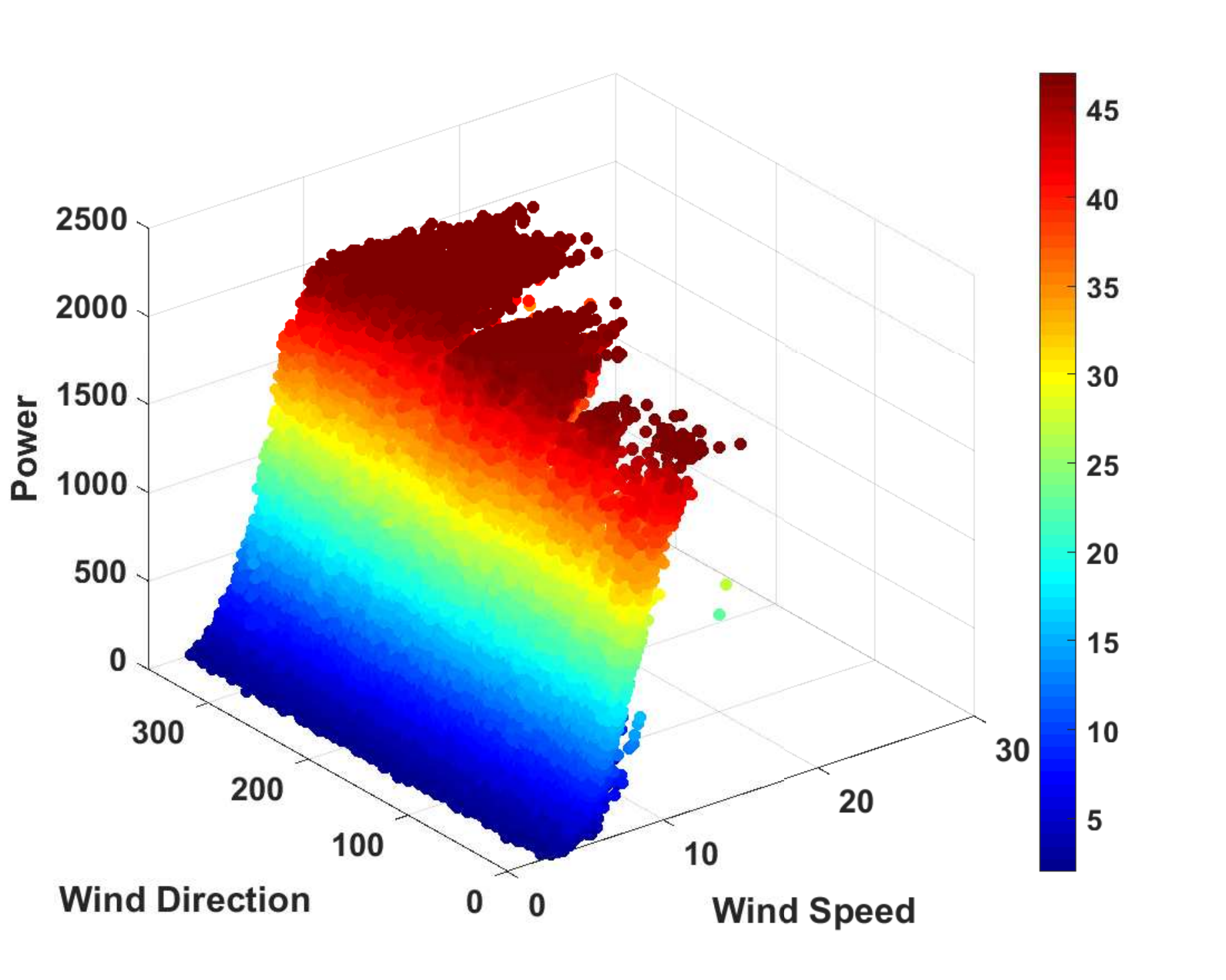}}
\subfloat[]{
\includegraphics[clip,width=0.31\columnwidth]{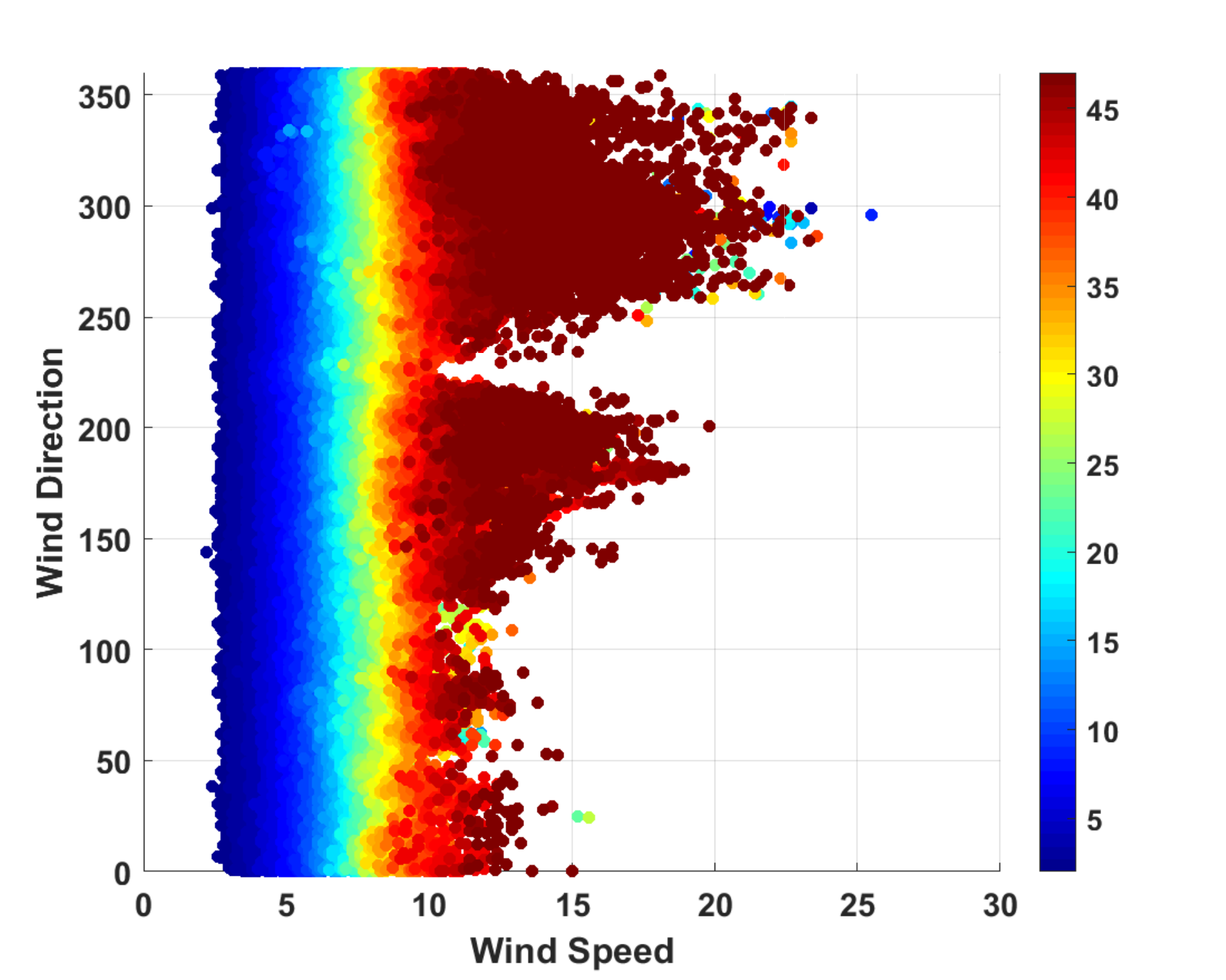}}
 \subfloat[]{
\includegraphics[clip,width=0.31\columnwidth]{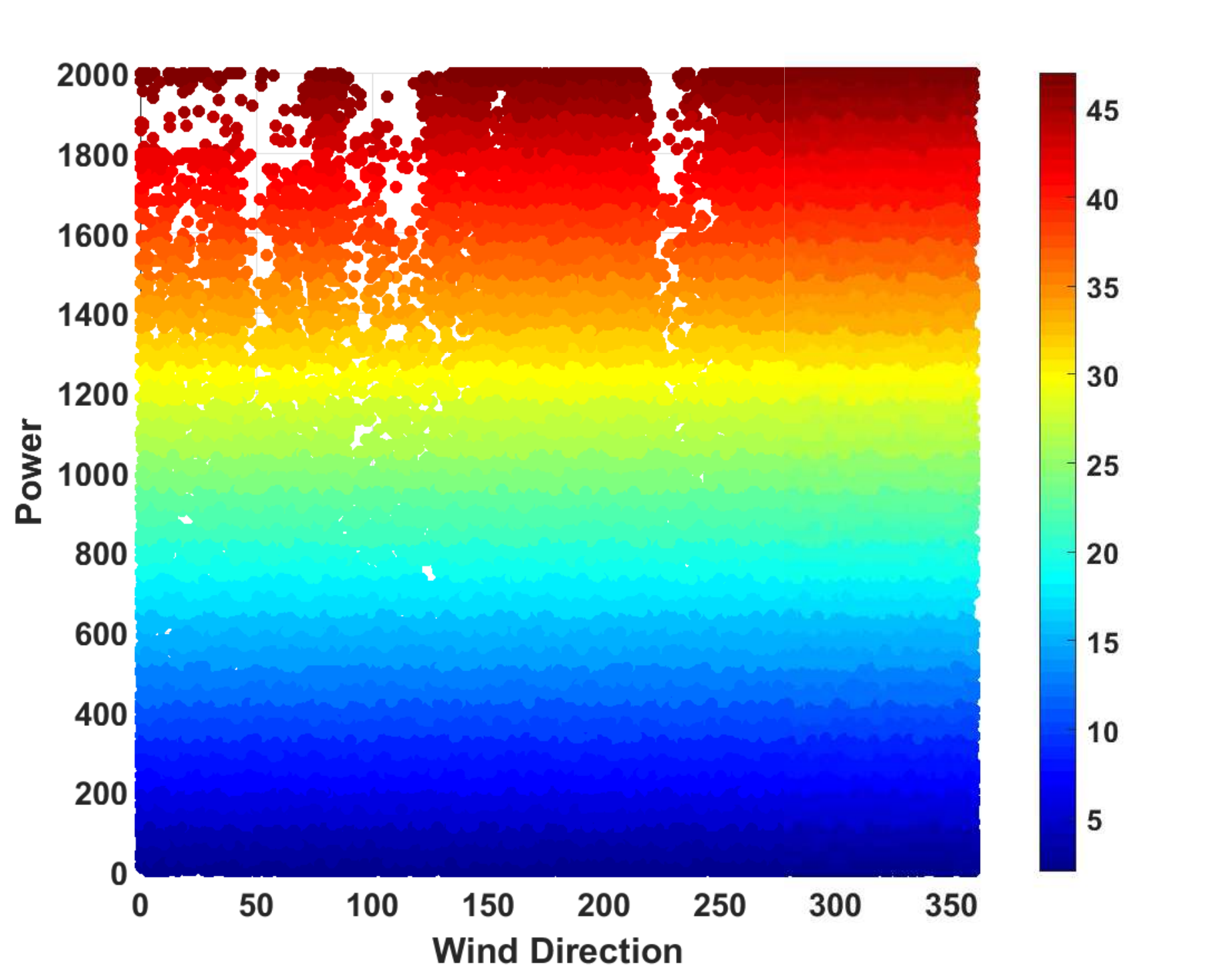}}
   \caption{The correlation between wind turbine power output and wind speed over the 42 months of data collection.(a) Outliers can be seen clearly in the data. (b) 3D figure of power curves, wind speed and wind direction. (c) the correlation between wind speed and direction. (d) power curves and wind direction.  }
   \label{fig:power_curve}
 
  \end{figure*}
 
 \begin{figure*}[tbp]
 \centering
  \includegraphics[width=0.6\textwidth]{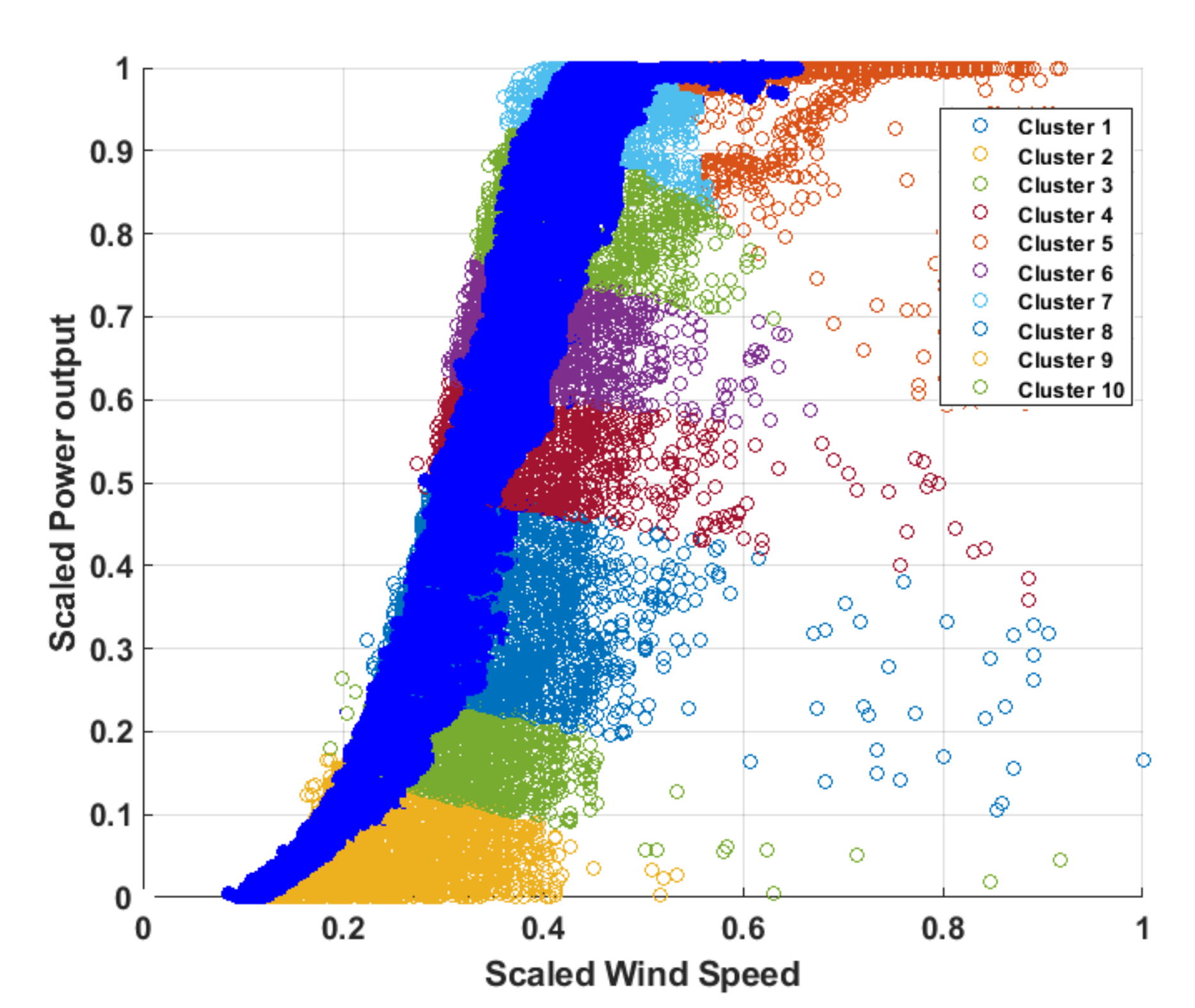}  \caption{ Clustering the data into 10 groups by K-means and then detecting and removing the outliers by an autodecoder NN. The purified data after removing outliers show by the dark blue region.   }
   \label{fig:outliers_detection}
 
  \end{figure*}

\section {Information preprocessing} \label{sec:preprocessing}
In data science, outliers are values that differ from regular observations in a dataset. 
Figure \ref{fig:power_curve} shows the correlation between wind turbine power output and wind speed and wind direction during the 42-month data collection period. The outliers can be seen in the scatter plots clearly which are distributed in the right side of the plot.  
In this study, we apply a combination of a K-means method which is one of the well-known Clustering Based Outlier Detection (CBOD)~\cite{jiang2001two}  methods and an autoencoder neural network to detect and remove the outliers from the SCADA dataset. 
As previous studies in Section~\ref{sec:Related-works} show,  wind speed is the primary factor that determines wind power from the SCADA features.  In the data, wind speed is widely distributed; we use the K-Means clustering algorithm to classify the wind power data into K subclasses. Before the clustering, due to the significant differences in numerical values of each type of data which has a significant impact on the training of the autoencoder's latent model, the data is normalized between zero and one~\cite{Qiuyi2019wind}. The normalization used in this paper is described in Equation~\ref{eq:normilization}.
\begin{equation}\label{eq:normilization}
\hat{Z}=\frac{Z-Z_{min}}{Z_{max}-Z_{min}}
\end{equation}

For this work, the number of the clusters is set to 10~\cite{Qiuyi2019wind}, which is applied for wind speed and power output of SCADA data. These clusters indicate the different operation states of the main subsystems, like the drive train and the control system~\cite{cui2018anomaly}.
Figure~\ref{fig:outliers_detection} shows ten clusters of Turbine 6. It can be observed that the distribution in each cluster is a horizontal band.
Within these bands, outliers are more easily discerned as being relatively far from the main body of the cluster. In order to remove the outliers in each cluster, an autoencoder neural network is used that shows better performance compared with other traditional outliers detection methods~\cite{an2015variational,moeini2017comparing}. An autoencoder is a particular type of unsupervised feedforward neural network which is trained to reconstruct output similar to each input. In this work the autoencoder consists of an input layer and one hidden layer which are fully connected~\cite{bengio2013representation}. 
For this goal, the input data are mapped to the hidden layer (encoding part), which typically comprises fewer nodes than the input layer and consequently compresses the data. Next, from the hidden layer, the reconstructed data flows through the output layer, which is re-transformed (decoding part), and the squared restoration error between the network's output and its input is calculated. For detecting the outliers, It is noticed that outliers have higher reconstruction error than the norm of the dataset. Therefore we remove the observations which have higher RMSE than the average of all data RMSE. Figure~\ref{fig:outliers_detection} presents the outliers detection and removal process, and the dark blue shows the clean data. 

\subsection{Performance criteria of forecasting models}
\label{sec:performance}
For evaluating and comparing the performance of the applied forecasting models, four broad performance indices are used: the mean square error (MSE), the root mean square error (RMSE), mean absolute error (MAE), and the Pearson correlation coefficient (R) \cite{zhang2017compound}. 
The equations for MAE, RMSE and R are described as follows :
\begin{equation}\label{eq:MAE}
\text{MAE}=\frac{1}{N}\sum_{i=1}^{N}|f_p(i)-f_o(i)|
\end{equation}
\begin{equation}\label{eq:RMSE}
\text{RMSE}=\sqrt{\frac{1}{N}\sum_{i=1}^{N}(f_p(i)-f_o(i))^2}
\end{equation}

\begin{equation}\label{eq:R}
R=\frac{\frac{1}{N}\sum_{i=1}^{N}(f_p(i)-\overline{f}_p)(f_o(i)-\overline{f}_o)}
{\sqrt{\frac{1}{N}\sum_{i=1}^{N}(f_p(i)-\overline{f}_p)^2}\times\sqrt{\frac{1}{N}\sum_{i=1}^{N}(f_o(i)-\overline{f}_o)^2}}
\end{equation}
where $f_p(i)$ and $f_o(i)$ denote the predicted and observed SCADA values at the $i^{th}$ data point. The total number of observed data points in $N$. The variables $\overline{f}_p$ and $\overline{f}_o$ are the means of the predicted and perceived power measures, respectively. For developing the effectiveness of the predicted model, MSE, RMSE and MAE should be minimised, while R should be maximised.

\section{Methodology}\label{sec:Methodology}
In this section, we introduce the proposed methodologies and related concepts for short-term wind turbine power output forecasting, including LSTM network details, self-adaptive differential evolution (SaDE) and the hybrid  LSTM network and the SaDE algorithm.

 \subsection{Long short-term memory deep neural network (LSTM)}

\begin{figure*}[tbp]
 \centering
  \includegraphics[width=0.4\textwidth]{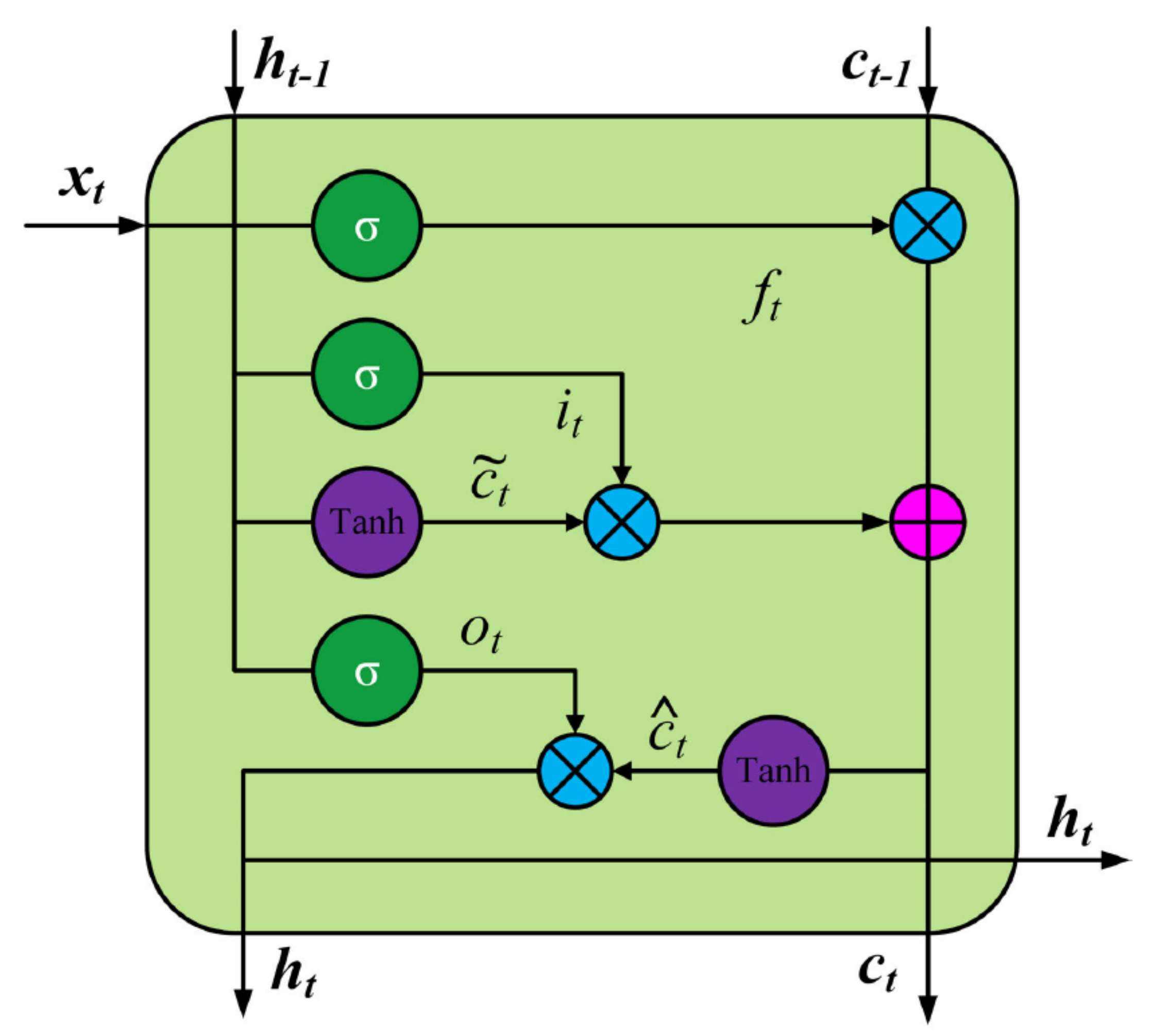}
   \caption{The internal structure of LSTM network from~\cite{zhou2019wind}.}
   \label{fig:LSTM}
 
  \end{figure*}

The LSTM network~\cite{hochreiter1997long} is a special kind of recurrent neural network (RNN) with three thresholds, namely the input gate, the output gate and the forgetting gate. The unit structure of the LSTM network can be seen in Figure \ref{fig:LSTM}. The forgetting gate defines the allowed rise or drop of the data flow~\cite{zhou2019wind} by placing the threshold, which indicates reservation and forgetting. Considering that an RNN hidden layer has only one state, there are severe difficulties with gradient fading and gradient explosion. Augmenting the RNN, the LSTM adds the structure of the cell state, which can recognise the long-term preservation of the state and emphasises the active memory function of the LSTM network. In the case of massive wind power time series data, the network can significantly enhance the accuracy of wind power prediction. In the forward propagation method of the LSTM network, the output value of the forgetting gate $f_t$ can prepare
the information trade-off of the unit state and the functional relationship encoded by Equation \ref{eq-LSTM1}.
\begin{equation}\label{eq-LSTM1}
f_t=\sigma (w_f h_{t-1}+u_f x_t+b_f)
\end{equation}
Both $i_t$ and $\tilde{c}_t$ are generated by the input gate,
which are related to the previous moment. The expressions
are as shown as Equation \ref{eq-LSTM2} and \ref{eq-LSTM3}.
\begin{equation}\label{eq-LSTM2}
i_t=\sigma (w_i h_{t-1}+u_i x_t+b_i)
\end{equation}
\begin{equation}\label{eq-LSTM3}
\tilde{c}_t=tanh (w_c h_{t-1}+u_c x_t+b_c)
\end{equation}
Cell state $c_t$ is the transmission centre of the cell state
before and later LSTM, which has the following functional
relationship:
\begin{equation}\label{eq-LSTM4}
c_t=c_{t-1} \odot f_t+i_t\odot \tilde{c}_t
\end{equation}
The output $h_t$ of the output gate derives from two components. the first part is the output of the previous moment that is the input of the
current moment and the second part is the information of the
current cell state and the particular expression model is delivered as Equation \ref{eq-LSTM5} and \ref{eq-LSTM6}.
\begin{equation}\label{eq-LSTM5}
o_t=\sigma (w_o h_{t-1}+u_o x_t+b_o)
\end{equation}
\begin{equation}\label{eq-LSTM6}
{h}_t=o_t \odot tanh (c_t)
\end{equation}
where both $u$ and $w$ are the weight values; $b$ and $\sigma$ are the bias values and activation function respectively, and $\odot$ is the Hadamard product.
For the LSTM network training settings, the Adam algorithm \cite{kingma2014adam} is employed to optimise the loss function, and  Dropout  \cite{gal2016theoretically} is used to prevent model overfitting.

\subsection{Self-adaptive Differential Evolution (SaDE)}
SaDE \cite{qin2005self} is proposed by Qin et al. to concurrently
perform two popular mutation strategies “DE/rand/1” and “DE/current-to-best/1.” SaDE adjusts the probability of generating offspring solutions using each strategy depending on the success rates (improved solutions) in the past $N_f$ generations of the algorithm. The aim of this adaptation scheme is to progressively evolve the best mutation strategy as search progresses.
This methodology is similar to the ideas proposed in~\cite{tvrdik2002competing}, where striving heuristics (including diverse DE variants, simplex methods and evolution strategies) are adopted simultaneously and probabilities for offspring generation are adjusted dynamically. 
 \begin{table} 
 \small
\caption{Summary of the best-found configuration for the predictive models tested in this paper (ten-minute ahead). }
\centering
\label{table:details_model}
\scalebox{0.8}{
\begin{tabular}{|l|p{9cm}|}
\hline 
 
\textbf{Models} & \textbf{Descriptions}\\ \hline\hline
 \textbf{ANFIS} \cite{pousinho2011hybrid}& Adaptive neuro-fuzzy inference system:
 \begin{itemize}
 \item  OptMethod= Backpropagation
   \item \textbf{Training settings}

  \begin{itemize}
 
\item  \textit{ErrorGoal}=0;
\item  \textit{InitialStepSize}=0.01;
\item  \textit{StepSizeDecrease}=0.9;
\item  \textit{StepSizeIncrease}=1.1;
\end{itemize} 
 
 \item \textbf{FIS features}
  \begin{itemize}
 
\item  \textit{mf} number=5;
\item  \textit{mf} type='gaussmf';

\end{itemize} 
 
 \end{itemize}
    \\  \hline
     \textbf{LSTM} \cite{hu2018nonlinear} + grid search& Long Short-term memory Network:
 \begin{itemize}
 \item \textbf{LSTM hyper-parameters}
 
 \begin {itemize}
 
 \item \textit{miniBatchSize}=512
 \item \textit{LearningRate}= $10^{-3}$
 \item \textit{numHiddenUnits1}   = 100;
\item \textit{Optimizer}= 'adam'

\end{itemize}

 \end{itemize}
 \\ \hline
\textbf{ CMAES-LSTM}~\cite{neshat2020evolutionary} &  
 
 \begin{itemize}
 \item \textbf{CMAES-LSTM hyper-parameters (Best configuration)}
 
 \begin {itemize}
 
 \item \textit{miniBatchSize}=1114
 \item \textit{LearningRate}= $10^{-4}$
 \item \textit{numHiddenUnits1}=201 ;
\item \textit{numHiddenUnits2}=30 ;
\item \textit{Optimizer}= 'adam'
\end{itemize}
 \end{itemize}
 \\ \hline
  \textbf{DE-LSTM}~\cite{peng2018effective} &  
 
 \begin{itemize}
 \item \textbf{DE-LSTM hyper-parameters (Best configuration)}
  \begin {itemize}
  \item \textit{miniBatchSize}=1155
 \item \textit{LearningRate}=$2.2\times10^{-3}$
 \item \textit{numHiddenUnits1}=141 ;
\item \textit{numHiddenUnits2}=42 ;
\item \textit{Optimizer}= 'adam'
\end{itemize}
 \end{itemize}
\\ \hline
\textbf{GWO-LSTM}~\cite{mirjalili2015effective,neshat2019adaptive} &  
 
 \begin{itemize}
 \item \textbf{GWO-LSTM hyper-parameters (Best configuration)}
  \begin {itemize}
  \item \textit{miniBatchSize}=1598
 \item \textit{LearningRate}=$0.3\times10^{-3}$
 \item \textit{numHiddenUnits1}= 150;
\item \textit{numHiddenUnits2}=235 ;
\item \textit{Optimizer}= 'rmsprop'
\end{itemize}
 \end{itemize}
\\ \hline
\textbf{SaDE-LSTM}&  
 
 \begin{itemize}
 \item \textbf{SaDE-LSTM hyper-parameters (Best configuration)}
  \begin {itemize}
  \item \textit{miniBatchSize}=727
 \item \textit{LearningRate}=$5.89\times10^{-3}$
 \item \textit{numHiddenUnits1}= 184;
\item \textit{numHiddenUnits2}=117 ;
\item \textit{Optimizer}= 'adam'
\end{itemize}
 \end{itemize}
 \\ \hline
\end{tabular}
}

\end{table}

In SaDE, the vector of the mutation factors are independently generated
at each iteration based on a normal distribution ($\mu=0.5$, $\sigma=0.3$), and trimmed to the interval (0, 2]. This scheme can retain both local (with small $F_i$ values) and global search capability to create potentially suitable mutation vectors during the evolution process. In addition, the crossover probabilities are randomly generated based on an independent normal distribution with $\mu=C_{Rm}$ and $\sigma=0.1$. This is in contrast to the $Fi$, the $C_{Ri}$ values which remain fixed for the last five generations before
the next regeneration. The $C_{Rm}$ is initially set to $0.5$.
For tuning $C_R$ to suitable values, the authors renew $C_{Rm}$ every
25 generations using the best $C_R$ values from  the last $C_{Rm}$ update.  

To speed up the SaDE convergence rate, a further local search procedure (quasi-Newton method) is used on some competent solutions after $N_s$ generations. The benefits of Self-adaptive parameter control make the SaDE as one of the most successful evolutionary algorithms, especially in real engineering optimisation problems that have multi-modal search spaces with many local optima~\cite{zhou2019self}.  

\subsection{Hybrid Neuro-Evolutionary Deep Learning method}
Multiple parameters for LSTM networks can influence their precision and performance. The selected hyperparameters include maximum training number of LSTM (Epoch), hidden layer size, the number of batch size, initial learning rate and the optimiser type.  If the maximum training number is too small, then the training data will be challenging to converge; if we set the number to a large value, then the training process might overfit.  The hidden layer size can influence the impact of the fitting~\cite{peng2018effective}. Batch size is also an important hyperparameter. If the batch size is set too small, then the training data will struggle to converge and will result in underfitting. If the batch size is too large, then the necessary memory will rise significantly. There are also complex interactions between hyperparameters. Therefore, a reliable optimisation technique should be utilised for tuning the optimal combination of hyperparameters to balance forecasting performance and computational efficiency.

There are three main methods for tuning hyper-parameters, including 1) manual trial and error,  2) systematic grid search, and 3) meta-heuristic approaches. In this paper, we apply the grid search and meta-heuristic approach, which is a self-adaptive version of DE (SaDE) for adjusting the optimal configuration of settings for the LSTM. This hybrid technique (SaDE-LSTM) is compared with the performance of  grid search; three hybrid neuro-evolutionary methods: DE-LSTM~\cite{peng2018effective}, CMAES-LSTM ~\cite{neshat2020evolutionary},  GWO-LSTM~\cite{mirjalili2015effective,neshat2019adaptive}; and ANFIS~\cite{pousinho2011hybrid}.

In the grid search method, we evaluate and tune just two hyperparameters of the LSTM: the batch size and the learning rate. Other settings assign a fixed value for the optimiser type, the number of LSTM hidden layers, the hidden layer size, maximum number of epochs by ('adam') \cite{kingma2014adam}, one, 100 and 100 respectively. These values are chosen for providing a baseline of the LSTM model evaluation.  
The ranges of batch size and learning rate are, respectively, selected from the ranges $128 \le BS \le 2048$ and $10^{-5} \le LR \le 10^{-1}$.

The optimization procedures are as follows:

\begin{itemize}
\item \textbf{Step 1}. Data preprocessing. Detecting and removing the outliers and then dividing the dataset into three subsets: the training, validation, and test sets.

  \item \textbf{Step 2}. Initialization. The parameters, maximum iteration number of SaDE, population size ($NP$), minimum and maximum crossover rate ($C_R$), mutation rate ($F$), and the upper and lower bounds of decision variables, the iteration numbers for updating the control parameters ( $N_f$ and $N_s$) are set. 
  
  \item \textbf{Step 3}. Generating offspring: The offspring solution is generated by the mutation, crossover, and selection operations, and is iterated until the offspring population is achieved.
  
  \item \textbf{Step 4}. Evaluating the offspring: The fitness values of the offspring population are computed by applying the proposed hyperparameters in the LSTM. The fitness is the root of the mean square error (RMSE) of the validation set; however, other performance indices are computed and recorded. The RMSE should be minimized, and the corresponding individual is the current best solution.   
  
  \item \textbf{Step 5}. Updating the SaDE control parameters by the historical optimization process. 
  
  \item \textbf{Step 6}. Stopping criteria:  if the maximum iteration is achieved, then SaDE is terminated, and the optimum configuration is taken; otherwise, the procedure returns to Step 3.
  
\end{itemize}

The fitness function of the optimisation process is defined in the following:

\begin{align}
\label{eq:fitness1}
\begin{split}
Argmin \to  f&=fitness(N_{h_1},N_{h_2},...,N_{h_D},N_{n_1h_1},N_{n_2h_2},...N_{n_Dh_D},L_R,B_S,Op) ,
\\
 Subject-to&:\\
 LN_h &\le N_h\le UN_h,\\
 LN_n &\le N_n\le UN_n,\\
10^{-5}&\le L_R \le 10^{-1},\\
128 &\le B_S \le 2048\\
1 &\le Op \le 3.
\end{split}
\end{align}
where $N_{h_i}, \{i=1,\ldots, D\}$ is the number of hidden layers for the $i-^{th}$ LSTM network and $N_{n_i,h_j}, \{j=1,\ldots, D_l\}$ is the number of neurons in the $i^{th}$ hidden  layer of this network. The lower and upper bounds of $N_h$ are presented by $LN_h$ and $UN_h$ , while $LN_n$ and  $UN_n$ are the lower and upper bounds of neuron number. The $Op$ is the selected optimizer for optimising the LSTM weights ('sqdm'~\cite{leen1994optimal}, 'adam'~\cite{kingma2014adam}, 'rmsprop'~\cite{tieleman2012lecture}).
\begin{figure}[tbp]
\centering
\subfloat[]{
\includegraphics[clip,width=0.47\columnwidth]{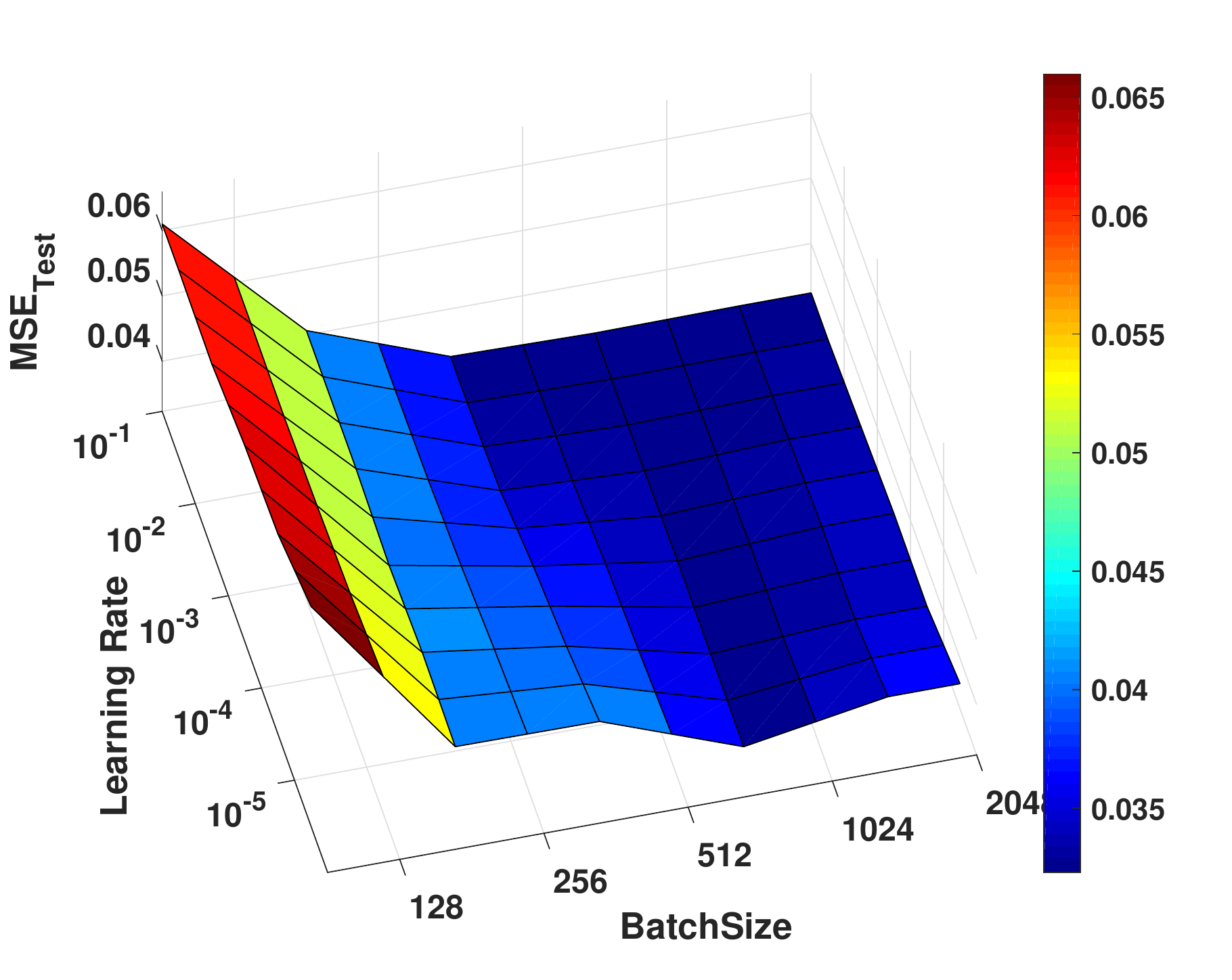}}
\subfloat[]{
\includegraphics[clip,width=0.47\columnwidth]{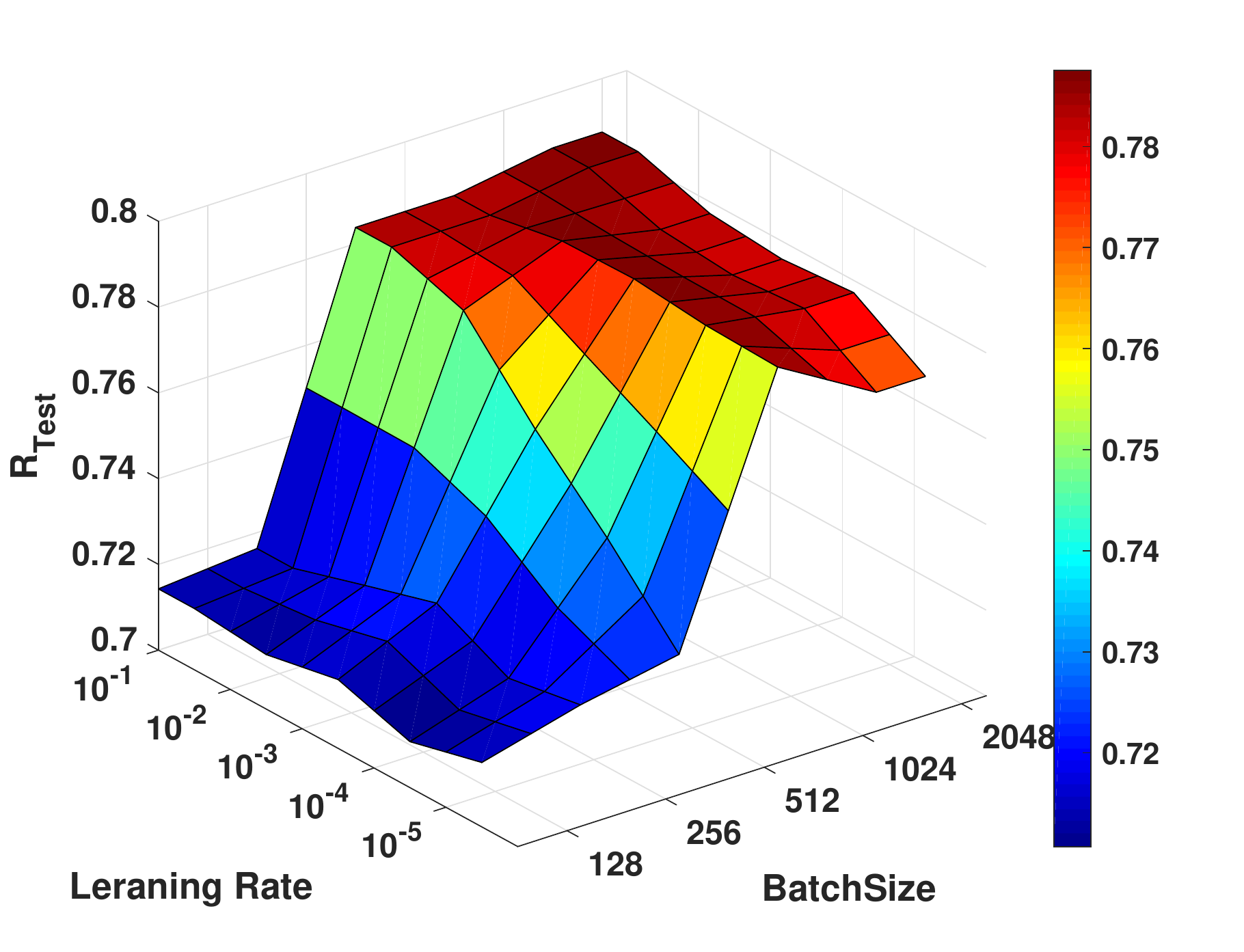}}\\
\subfloat[]{
\includegraphics[clip,width=0.47\columnwidth]{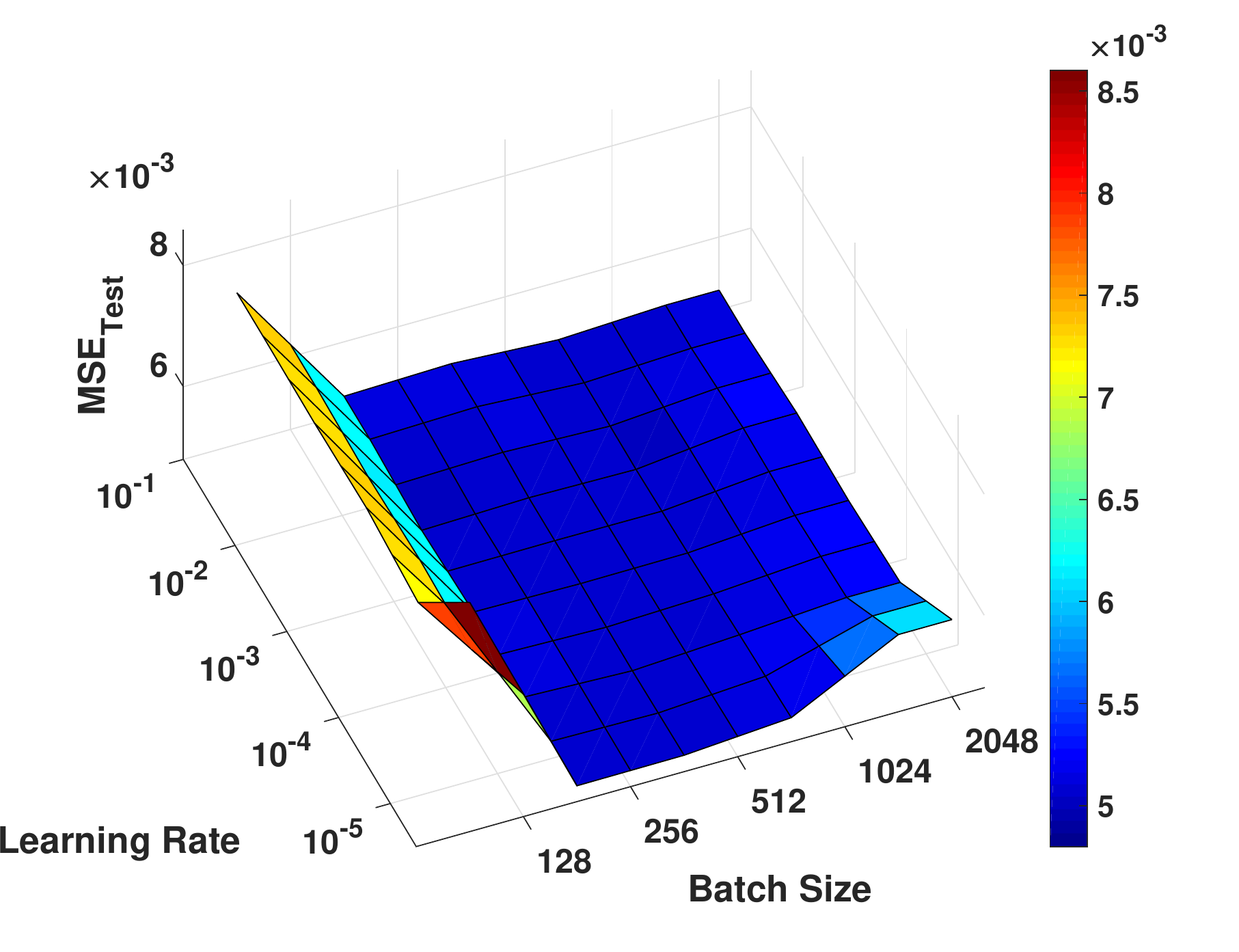}}
\subfloat[]{
\includegraphics[clip,width=0.47\columnwidth]{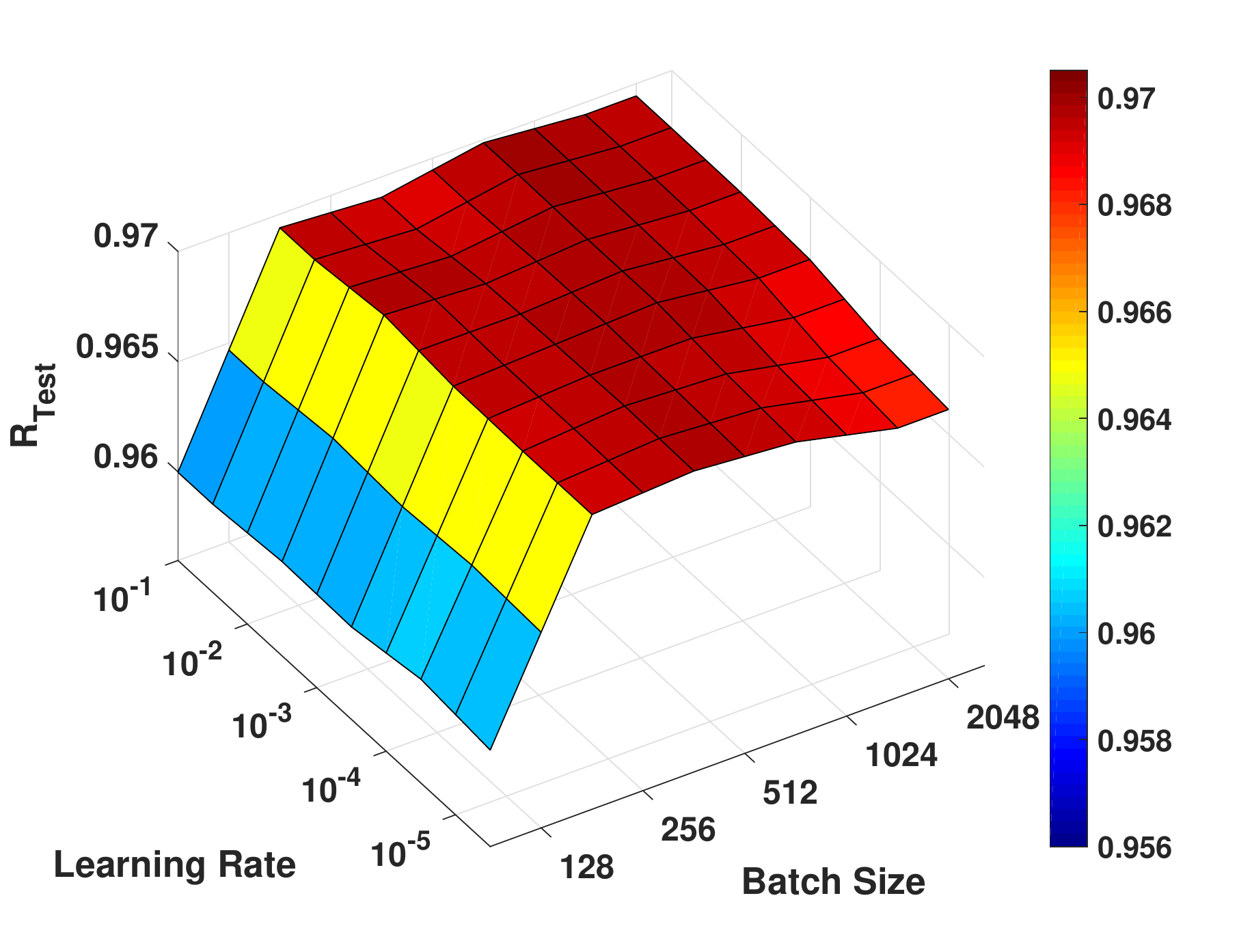}}\\
\subfloat[]{
\includegraphics[clip,width=0.47\columnwidth]{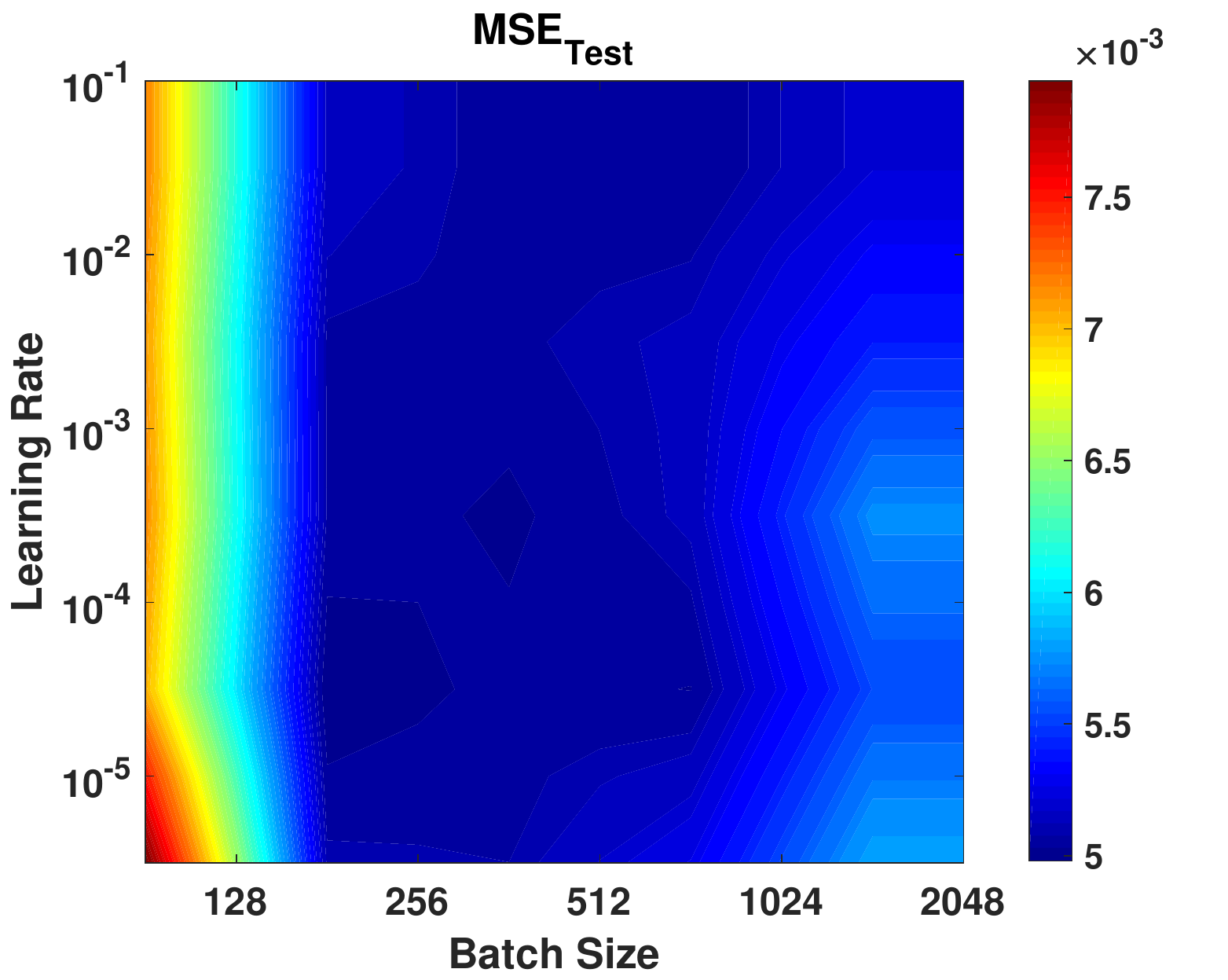}}
\subfloat[]{
\includegraphics[clip,width=0.47\columnwidth]{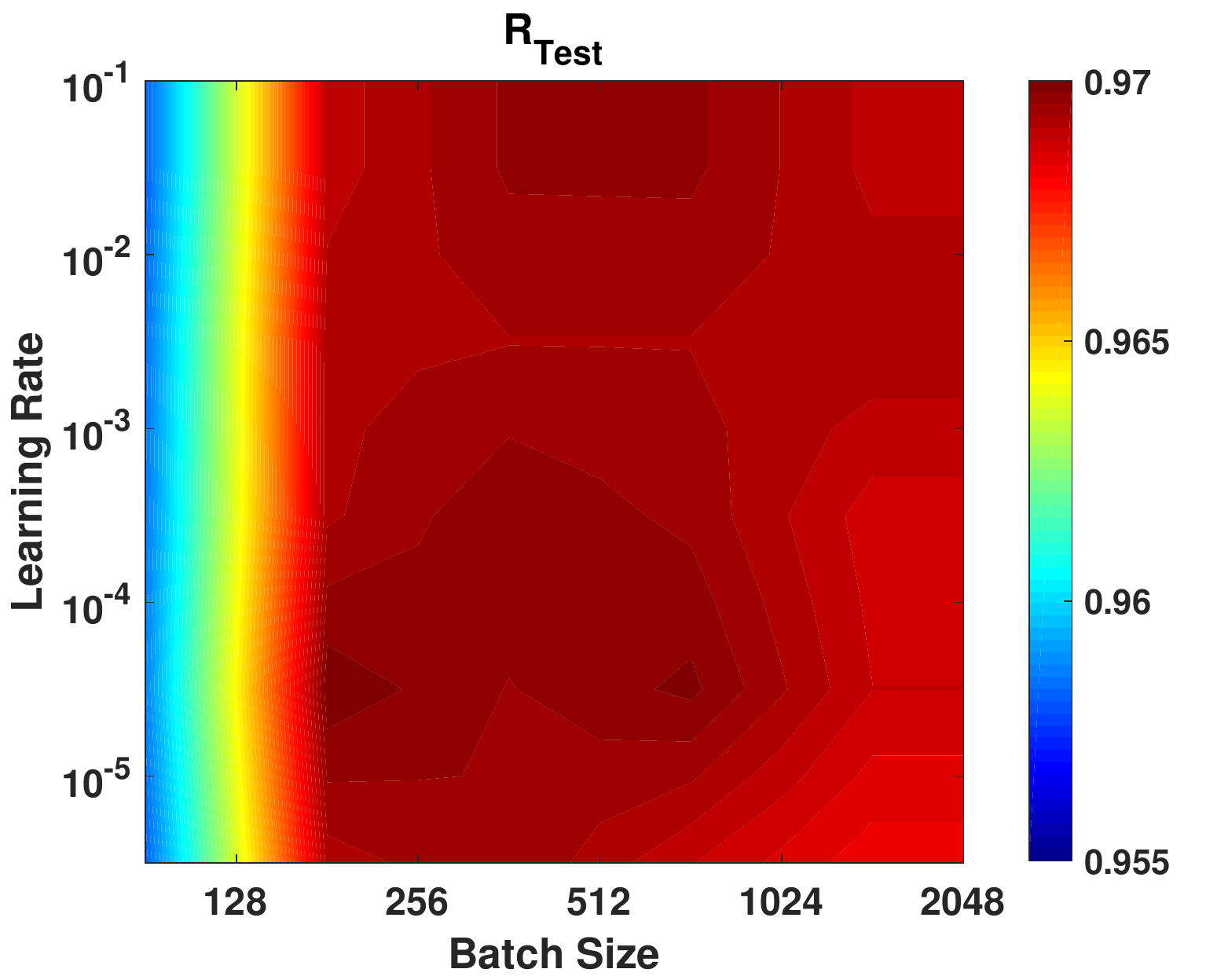}}
\caption{Hyper-parameter tuning of the applied LSTM network for forecasting the short-term power output of the wind turbine without removing the outliers(Layer number=1, neuron number=100, Optimizer='Adam') .(a) the average of MSE test-set (ten-minute ahead) with one input (wind speed)  (b) the average of R-value test-set (ten-minutes ahead) with one input (wind speed).(c) and (d) the average of MSE and R-value test-set  with two inputs (wind speed, Power) respectively. (e) and (f)the average of MSE and R-value test-set  with two inputs (wind speed, wind direction and Power) respectively }%
\label{fig:3Dplot_gridsearch_1}
\end{figure}
\begin{figure}[tbp]
    \centering
    \includegraphics[width=0.7\columnwidth]{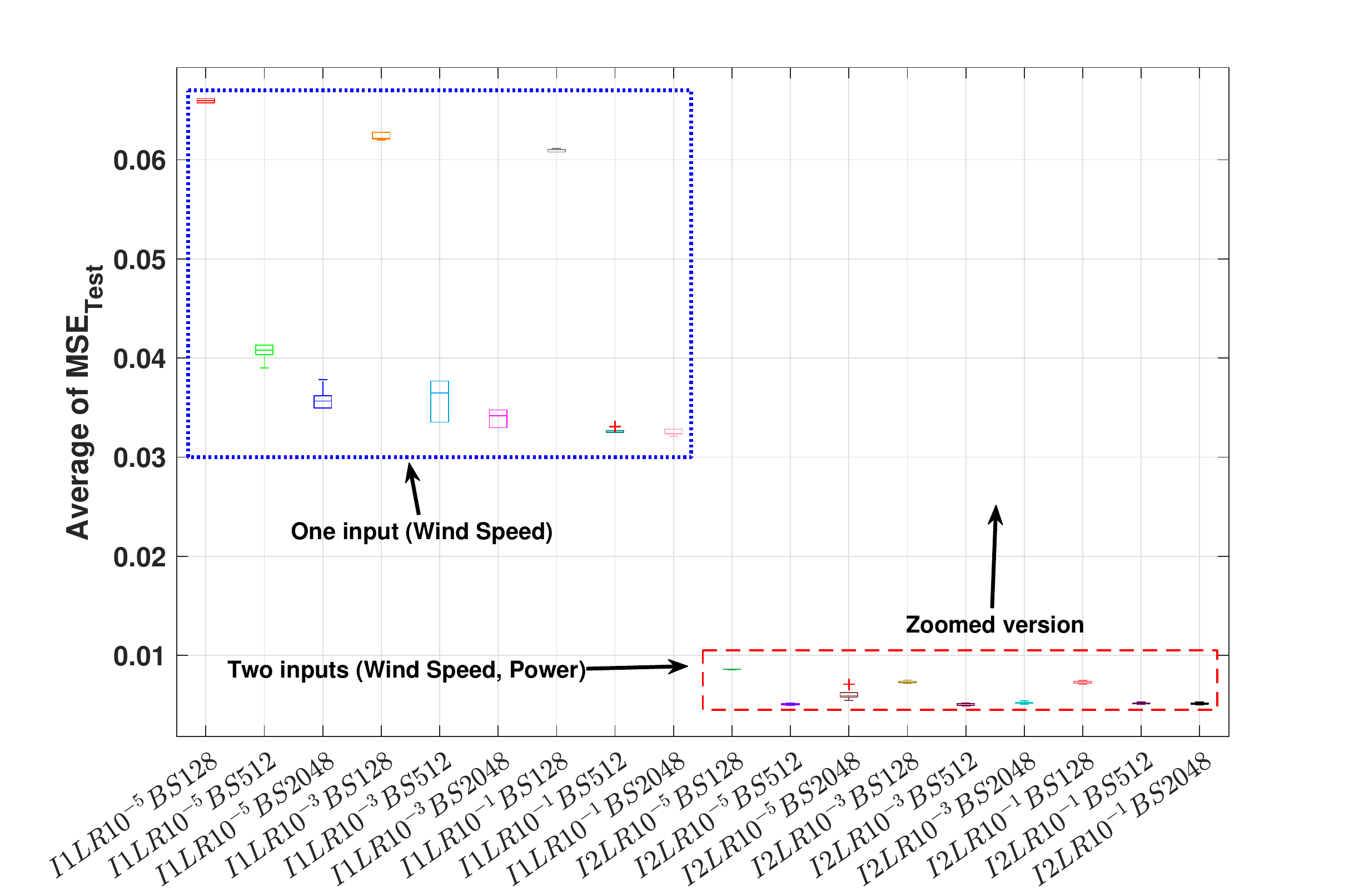}\llap{\raisebox{3.3cm}{\includegraphics[height=3.7cm]{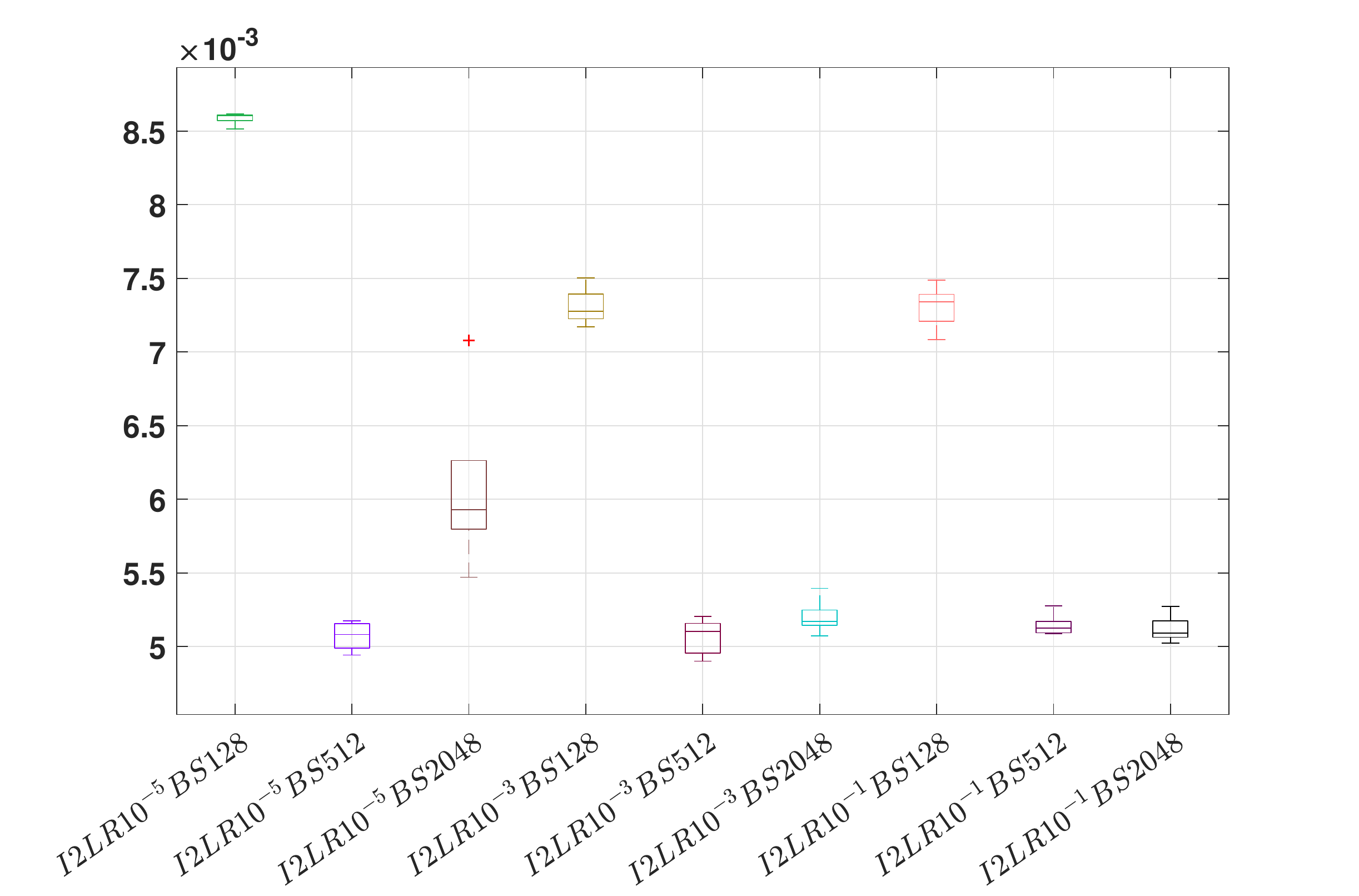}}}
     \caption{Comparison of the LSTM performance with one ($I_1$) and two inputs ($I_2$) without removing outliers }
    \label{fig:boxplot_twoinput_without}
\end{figure}
  \subsection{Experimentation design}
 In the first step of the forecasting power output of the wind turbine, we proposed four DNN models with different inputs and the same output. The main aim of proposing these forecasting models is analysing the impact of three SCADA features, wind speed, wind direction and the currently generated power on the predicting accuracy of the power output both ten-minutes and one-hour ahead. In the second step, we compare the performance of the proposed models before and after removing the outliers from the SCADA dataset to illustrate the effectiveness of the outlier detection technique (K-means + Autoencoder). Finally, the proposed hybrid model (SaDE-LSTM) is compared with some of the state-of-the-art forecasting frameworks.     

For evaluating the performance of four models with raw SCADA data which are categorised in three training ($80\%$), testing ($10\%$) and validating ($10\%$) sets randomly, the LSTM deep network is used, which is composed of one sequence input layer, one LSTM layer, fully-connected layer and a regression layer. A grid search method is used for tuning the hyperparameters, batch size and learning rate. Figure~\ref{fig:3Dplot_gridsearch_1} presents the performance of LSTM framework with a tuned batch size and learning rate parameters for three forecasting models in the interval of ten-minute.     The best performance of model 1 (one input) is obtained where the values of batch size are bigger than 512 and learning rate placed between the range of $10^{-2}$ and $10^{-4}$. The forecasting behaviours of both models 2 and 3 are similar, and the best accuracy is related to the batch sizes more than 256 and a learning rate between $10^{-3}$ and $10^{-5}$.


\begin{figure}[tbp]
\centering
\subfloat[]{
\includegraphics[width=0.49\columnwidth]{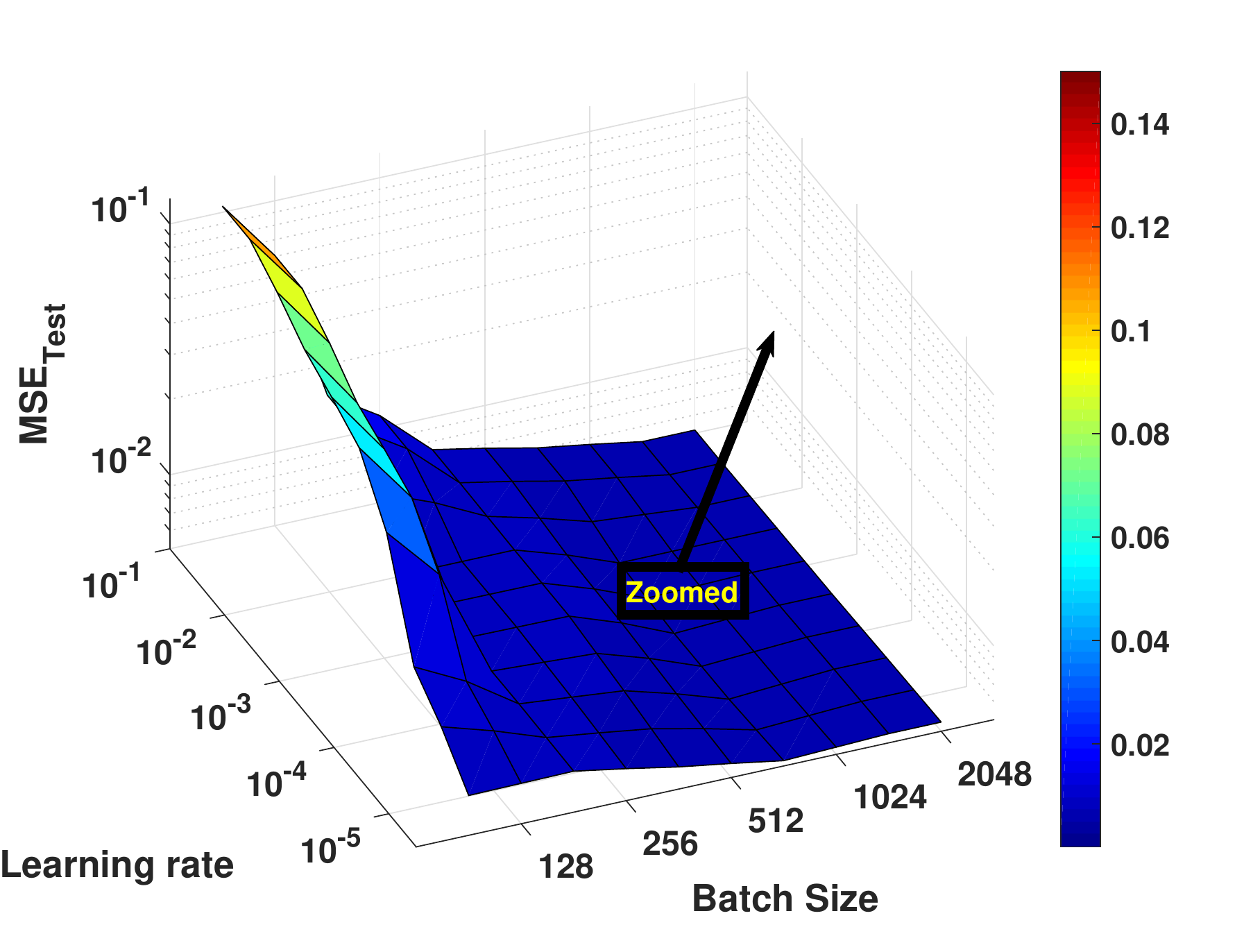}\llap{\raisebox{3.95cm}{\includegraphics[height=3cm]{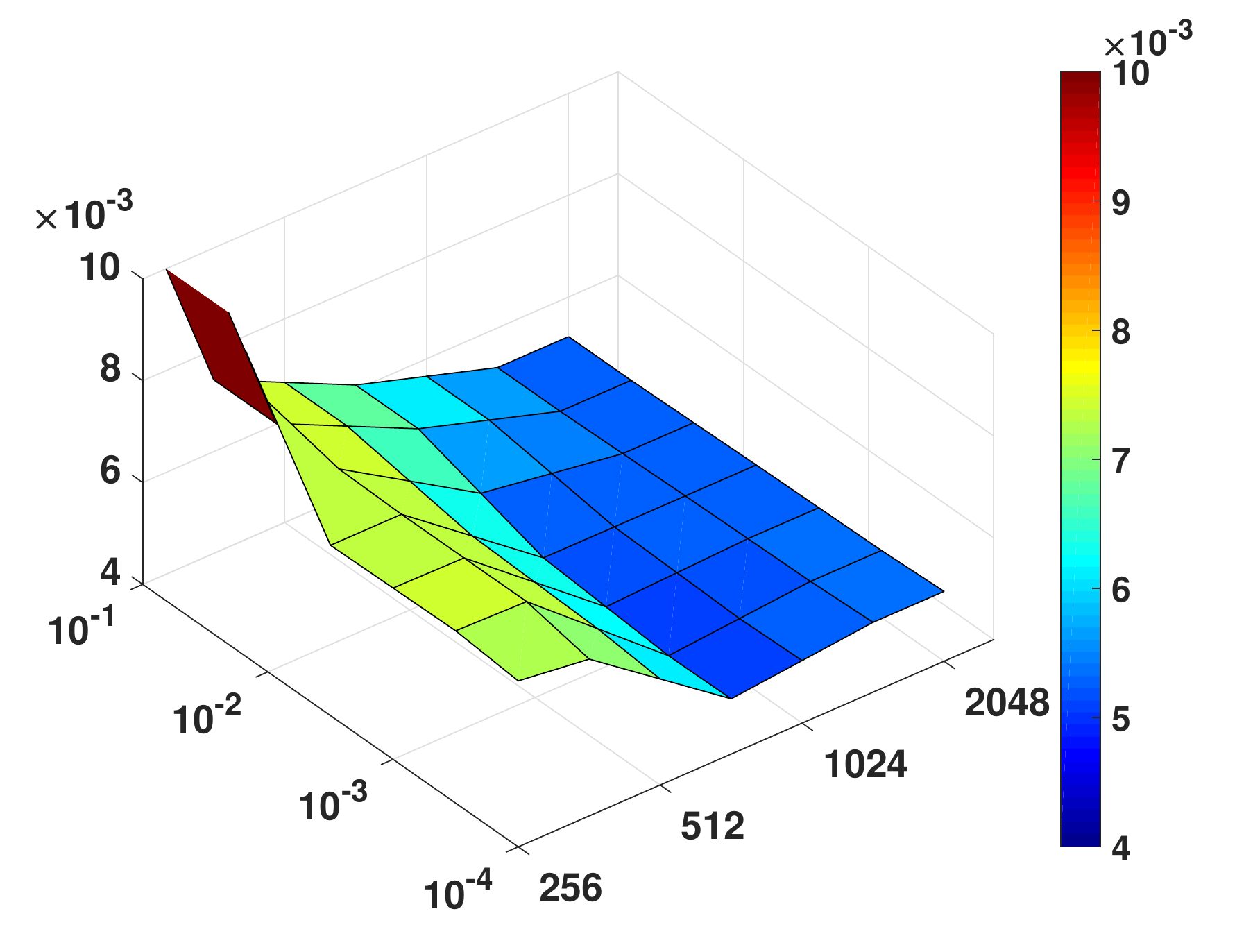}}}}
\subfloat[]{
\includegraphics[width=0.5\columnwidth]{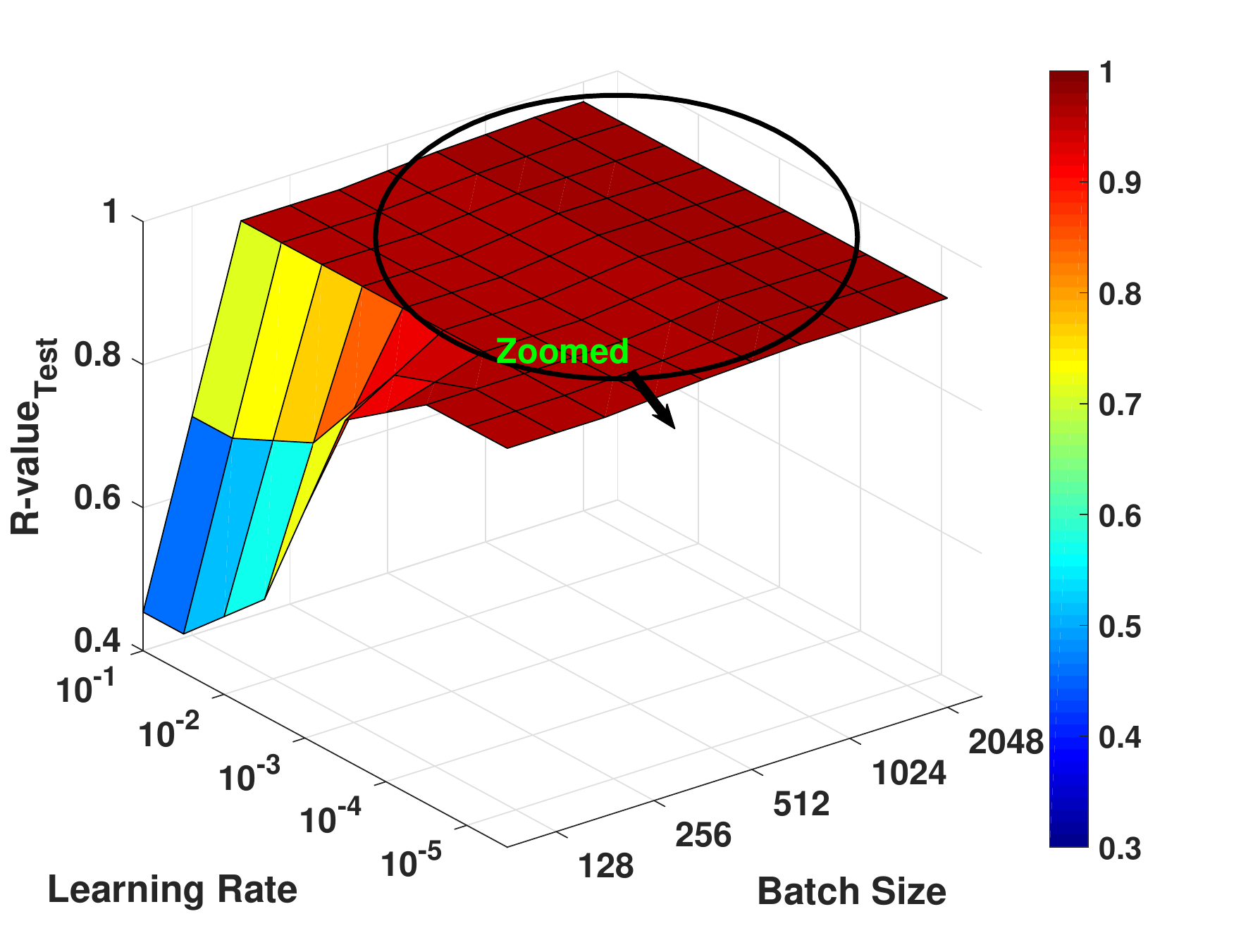}\llap{\raisebox{.5cm}{\includegraphics[height=3cm]{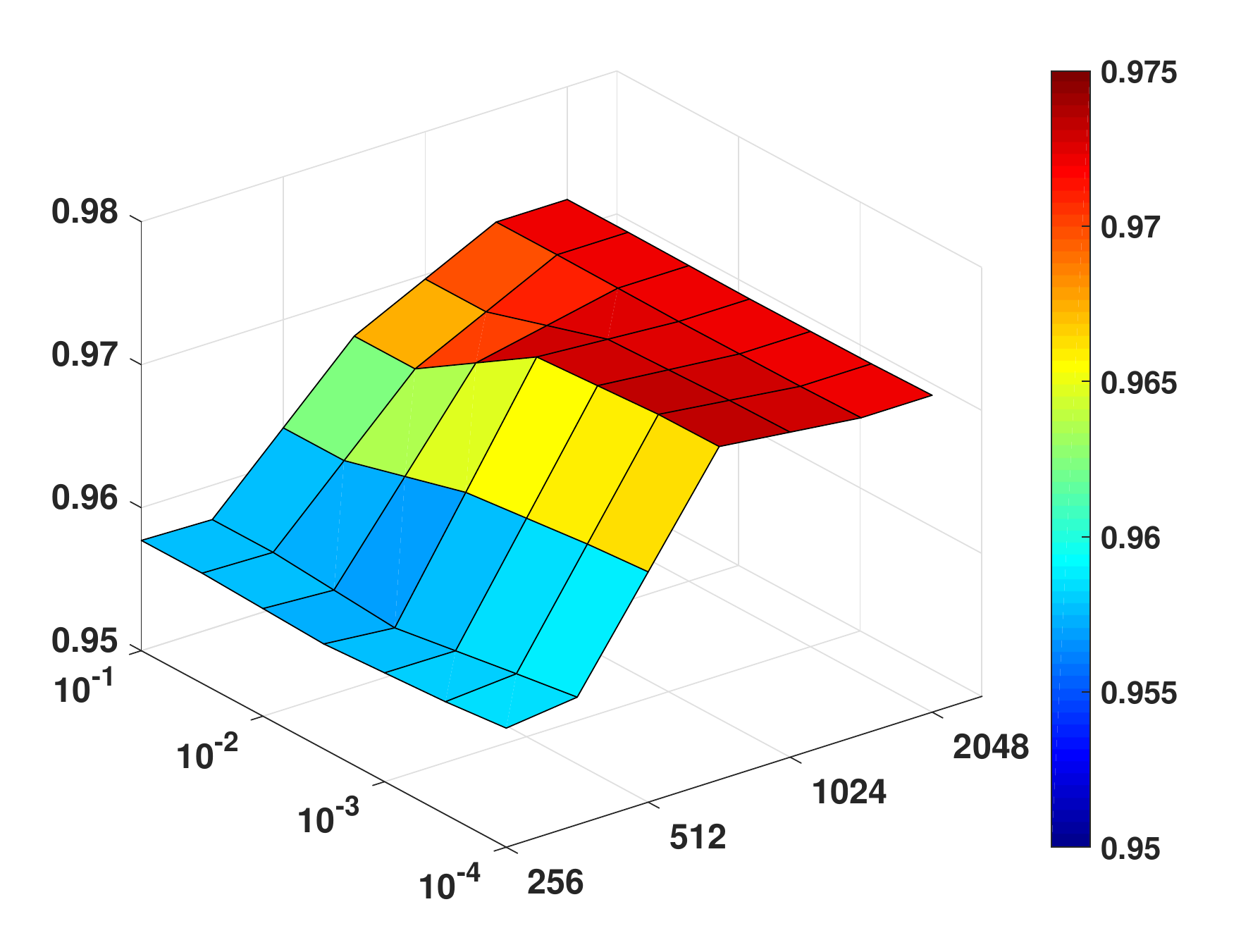}}}}

\caption{Hyper-parameter tuning of the applied LSTM network for forecasting the short-term power output of the wind turbine after removing outliers(Layer number=1, neuron number=100, Optimizer='Adam') .(a) the average of MSE test-set (ten-minute ahead) with one input (wind speed)  (b) the average of R-value test-set (one-hour ahead) with two inputs (wind speed and direction). }%
\label{fig:3Dplot_clean_ten}
\end{figure}

Figure \ref{fig:boxplot_twoinput_without} shows an average performance (MSE and R) comparison between two forecasting models 1 and 2 and shows that using the currently generated power as an input plays a significant role in producing an accurate prediction. 

After removing the outliers, we compare the performance of the LSTM framework for predicting the power produced by the $6^{th}$ wind turbine in two models (model 1 with one input and model 3 with two inputs including wind speed and direction) with different ranges of the batch size and learning rate. Figure~\ref{fig:3Dplot_clean_ten} illustrates the 3D forecasting landscape of the correlation between the batch size, learning rates and forecasting accuracy for model 1 (within ten-minute)  and model 3 (one-hour interval). The applied method for tuning the hyperparameters is the grid search in this experiment. In model 1, the best prediction results happen where the batch size is more than 1024 and the learning rate value is between $10^{-3}$ and $10^{-4}$.

\begin{figure}[t]
\centering
\subfloat[]{
\includegraphics[width=0.48\columnwidth]{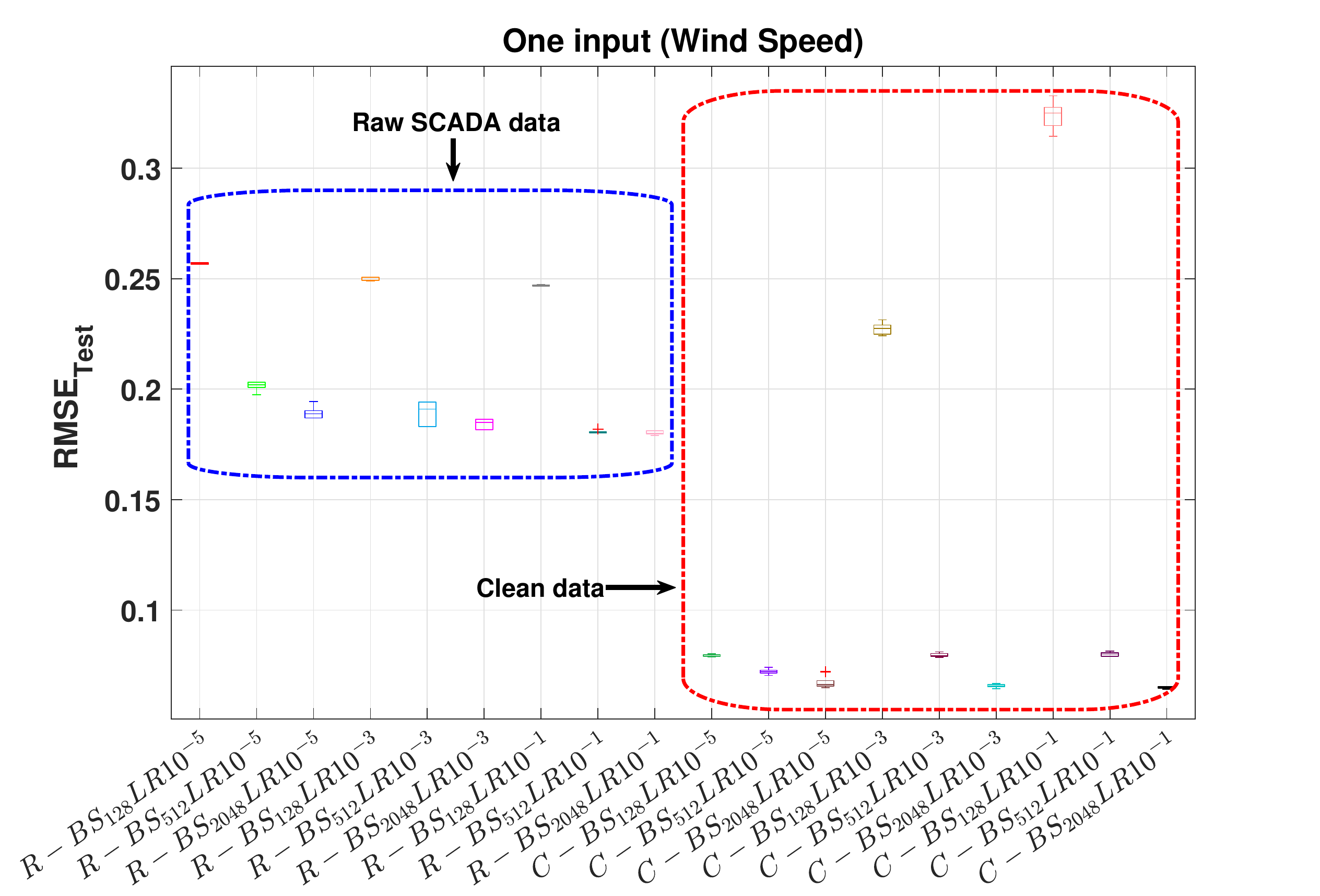}}
\subfloat[]{
\includegraphics[width=0.49\columnwidth]{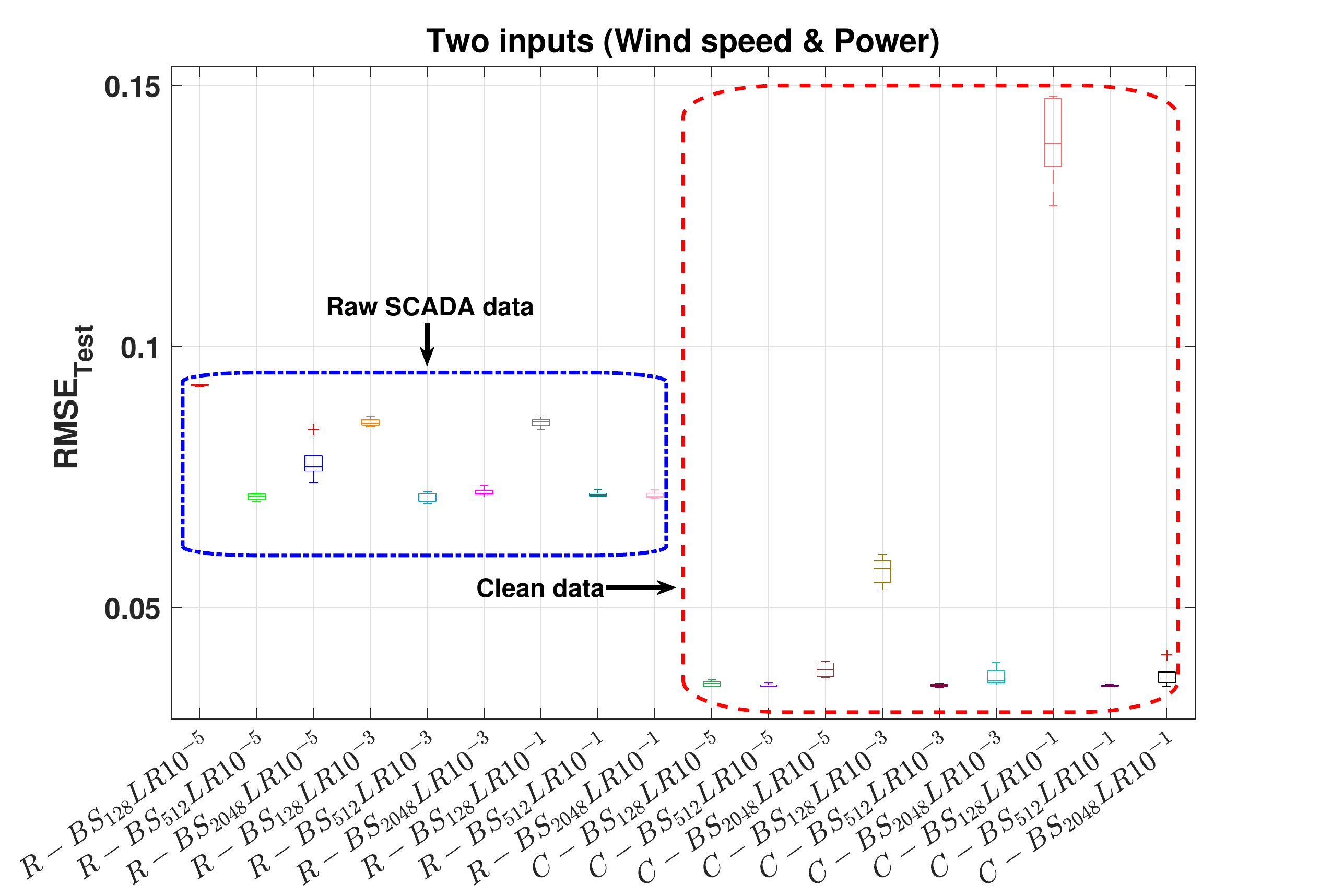}}\\
\subfloat[]{
 \includegraphics[clip,width=0.49\columnwidth]{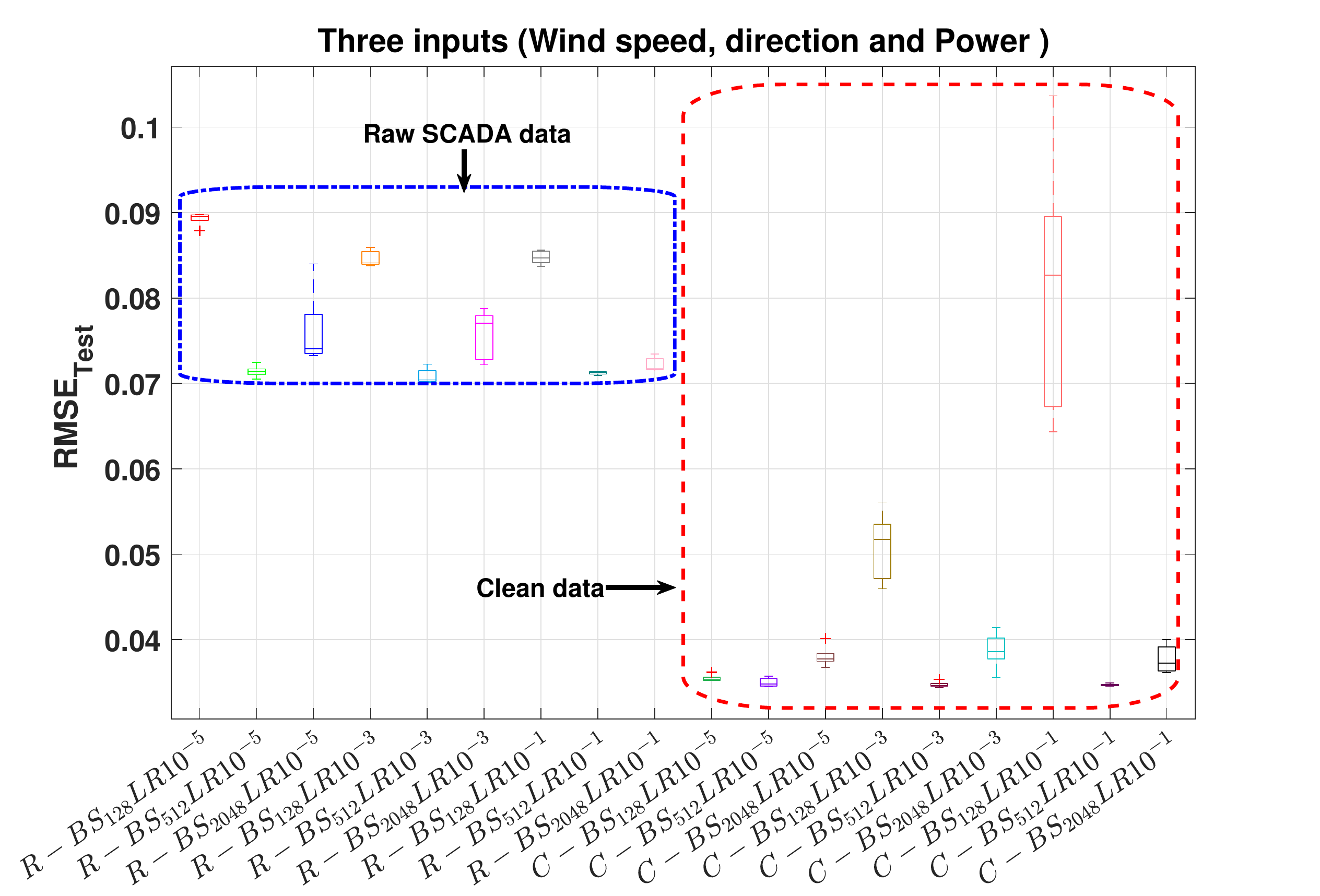}}
\subfloat[]{
\includegraphics[clip,width=0.49\columnwidth]{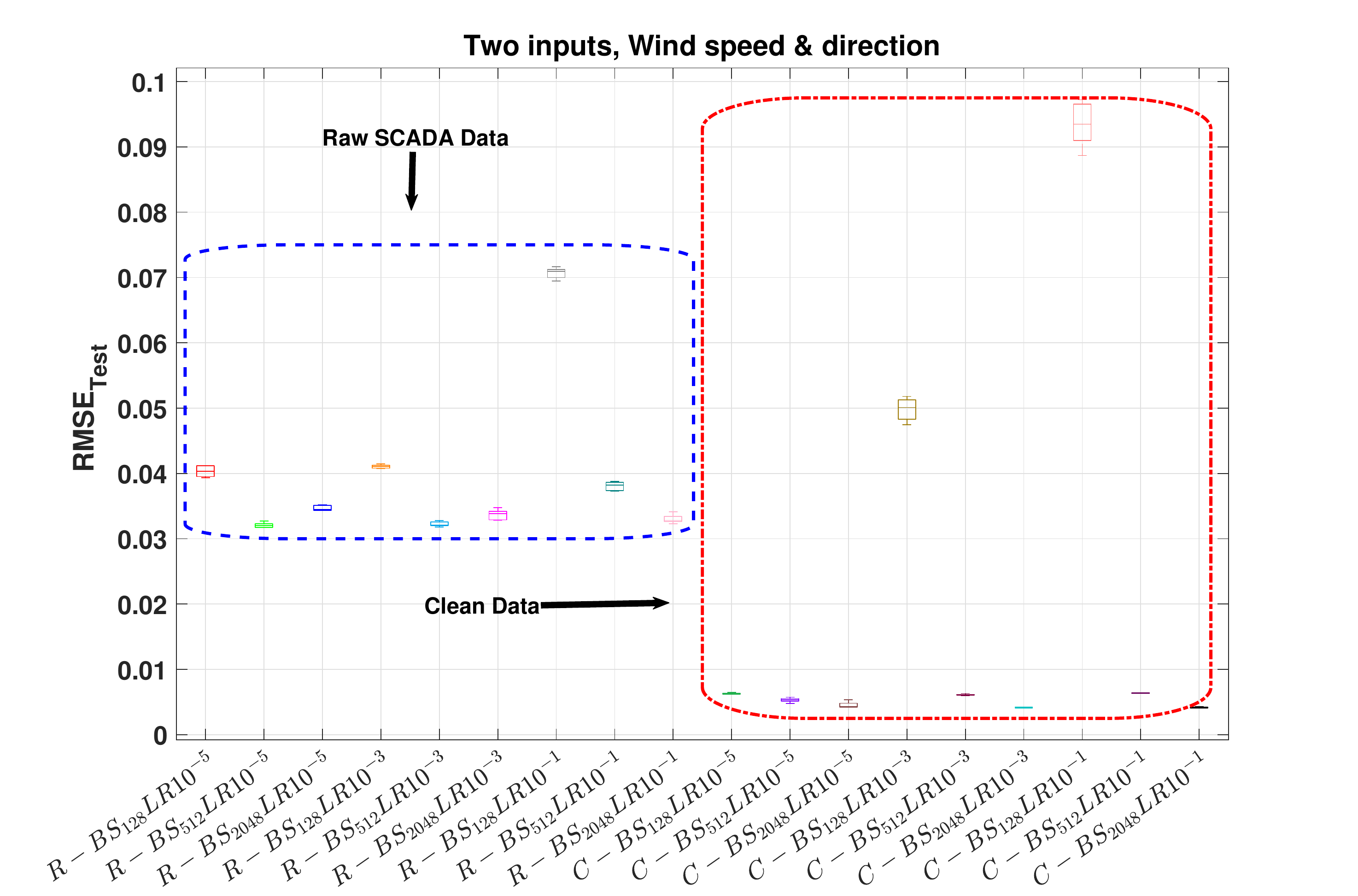}}

\caption{A comparison of the LSTM network Hyper-parameter tuning performance training on the raw SCADA data (R) and training after removing the outliers (C) for ten-minute ahead forecasting (Layer number=1, neuron number=100, Optimizer='Adam') .(a) the RMSE test-set with one input (wind speed)  (b) the RMSE test-set with two inputs (wind speed and current power). }%
\label{fig:box_compare_outliers}
\end{figure}
For highlighting the benefit of using the outlier detection $\&$ removal technique (K-means + Autoencoder), a comprehensive comparison for four prediction models is applied (the considered time step of the inputs  is ten-minutes), and the results of this experiment can be seen in Figure~\ref{fig:box_compare_outliers}.  Meanwhile, it is noticed that removing the noise from the SCADA data results in a significant enhancement in the accuracy of the prediction except two inappropriate configurations (Batch size=$128$ and Learning rate=$10^{-3}$, $10^{-1}$). 
\begin{figure}[t]
\centering
\subfloat[]{
\includegraphics[width=0.47\columnwidth]{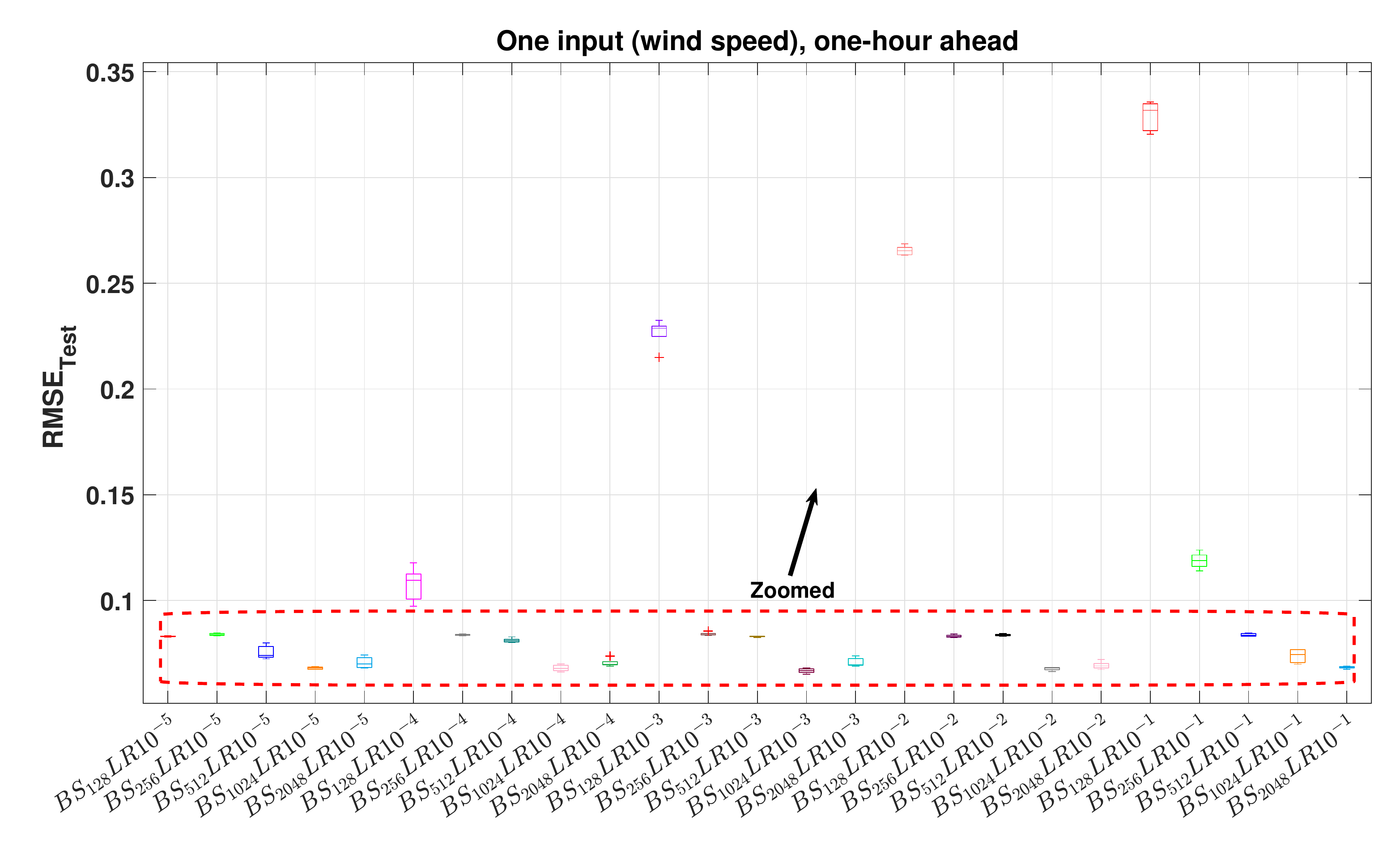}}\llap{\raisebox{1.5cm}{\includegraphics[height=2.5cm]{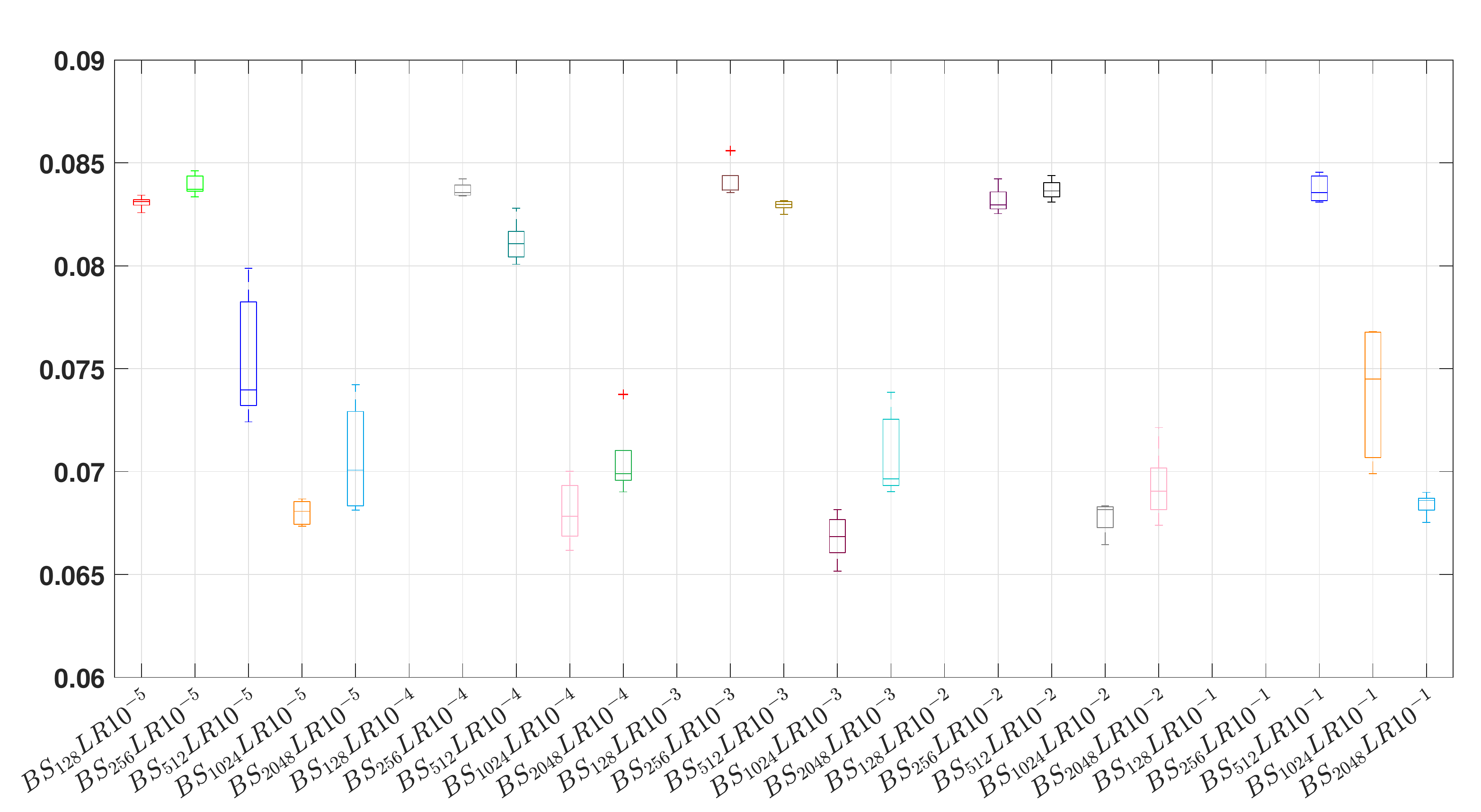}}}
\subfloat[]{
\includegraphics[width=0.47\columnwidth]{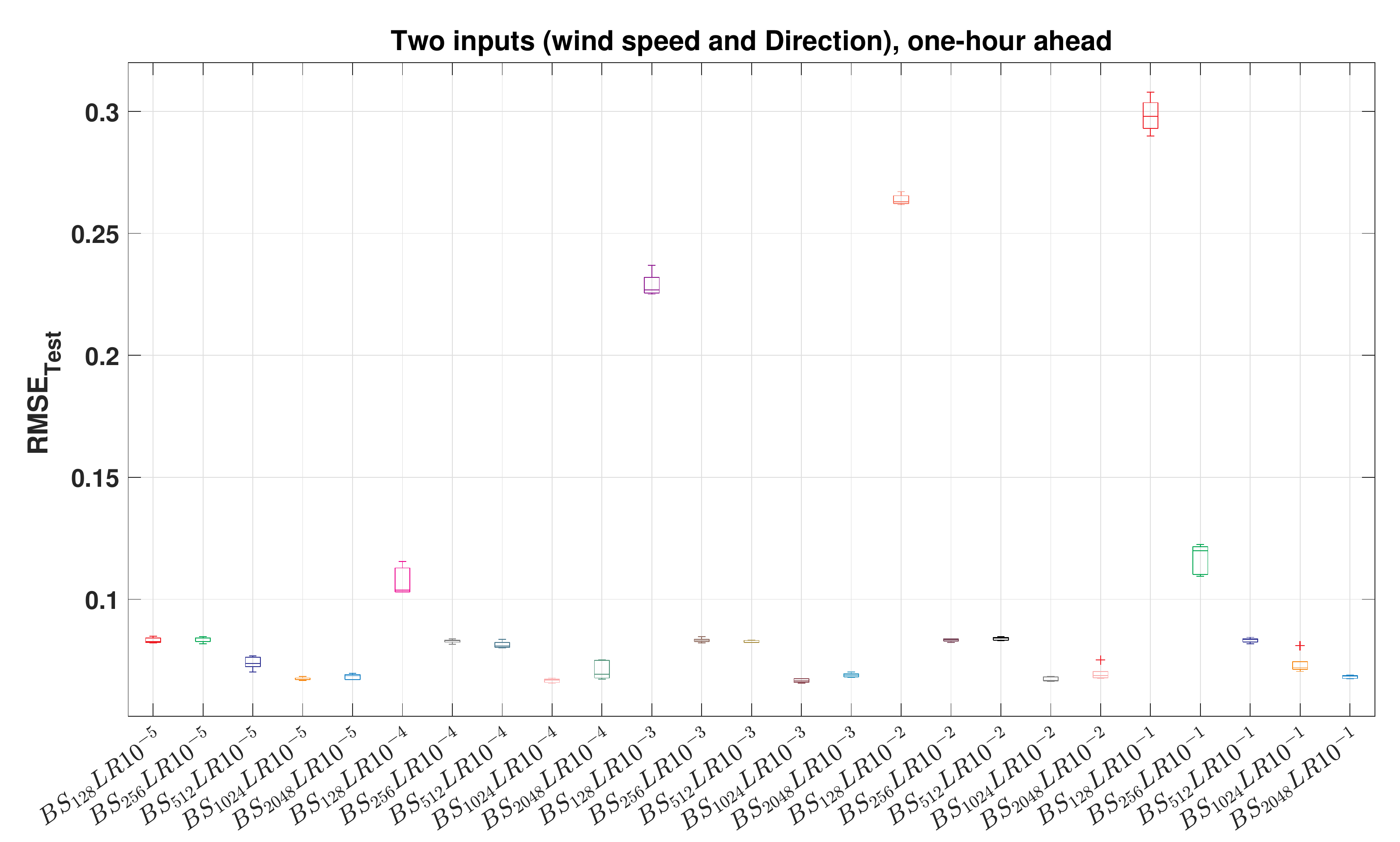}}\\
\subfloat[]{
 \includegraphics[clip,width=0.47\columnwidth]{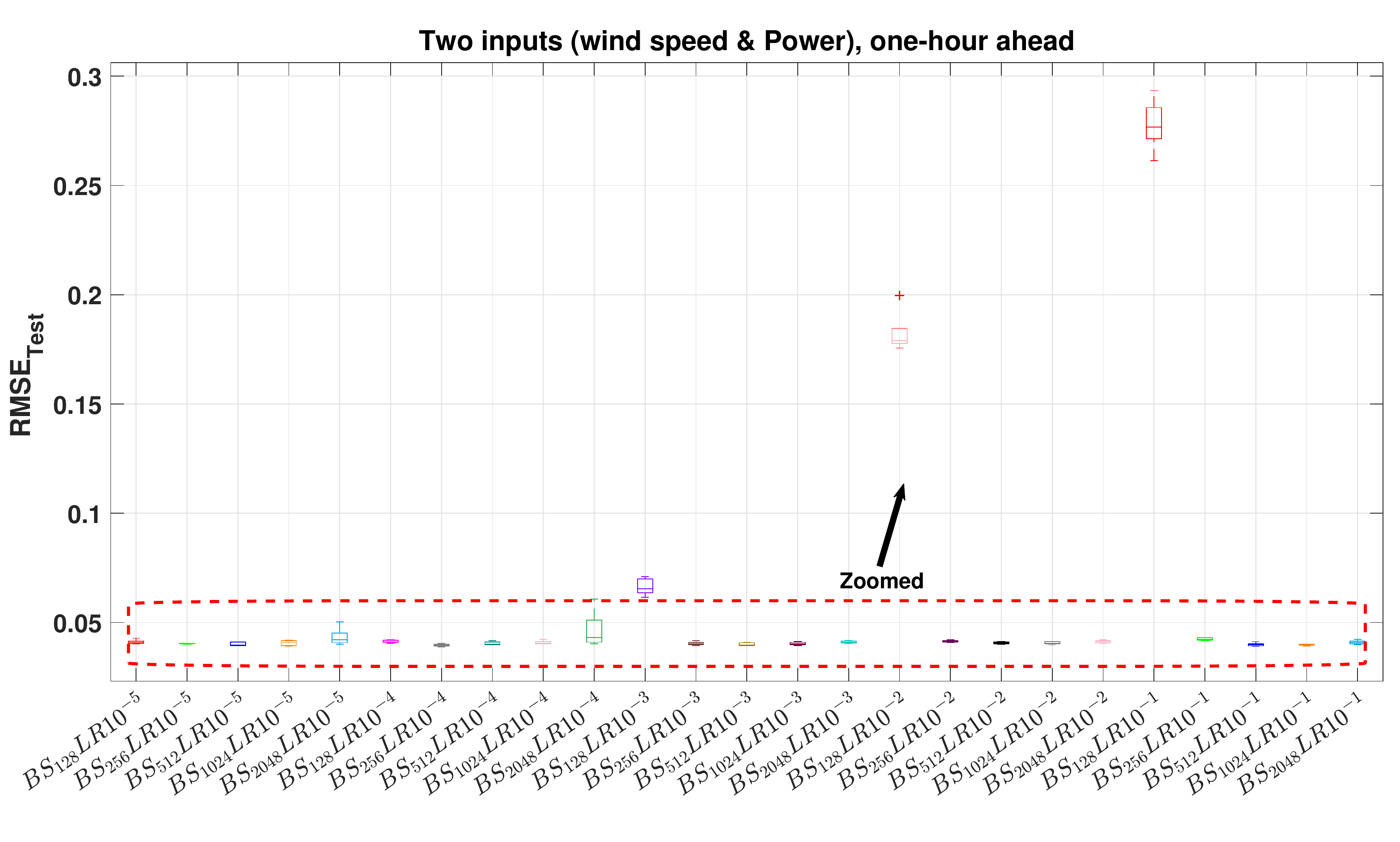}}\llap{\raisebox{1.55cm}{\includegraphics[height=2.5cm]{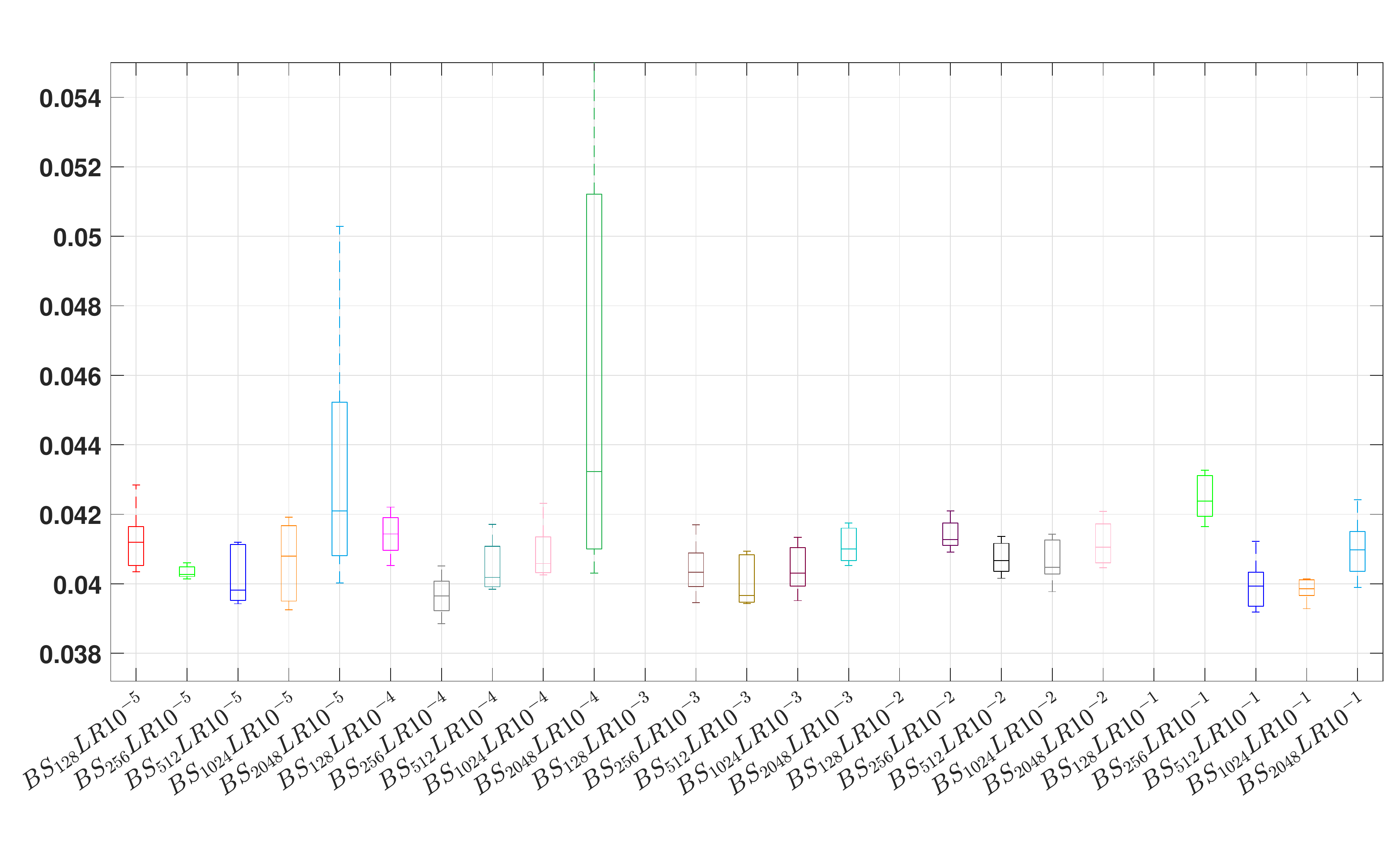}}}
\subfloat[]{
\includegraphics[clip,width=0.47\columnwidth]{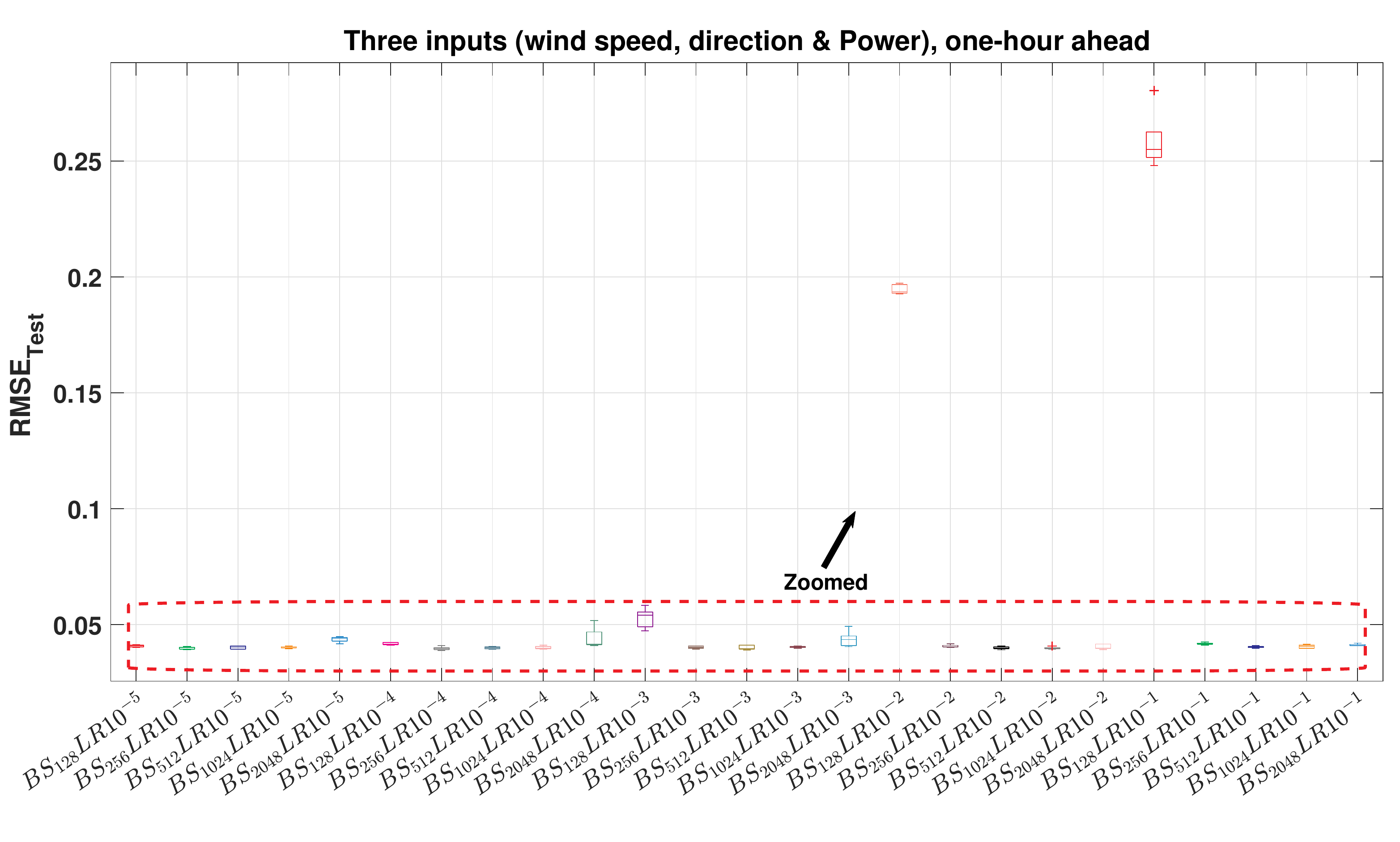}}\llap{\raisebox{1.55cm}{\includegraphics[height=2.6cm]{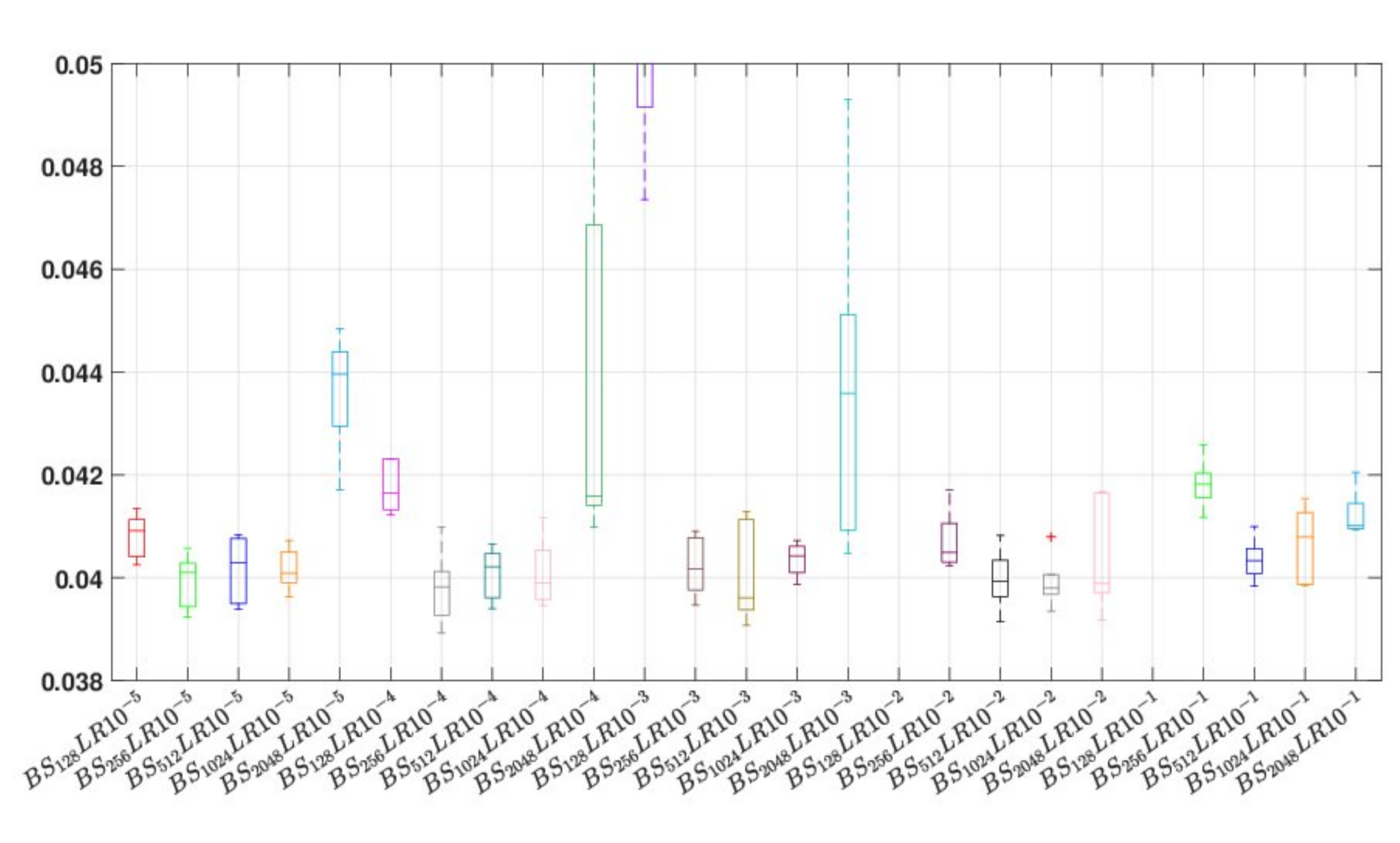}}}

\caption{A comparison of four forecasting LSTM network models performance with various  Hyper-parameters for forecasting the power output in one-hour ahead  (Layer number=1, neuron number=100, Optimizer='Adam') .(a) the RMSE test-set with one input (wind speed)  (b) the RMSE test-set with two inputs (wind speed and direction), (c) the RMSE test-set with two inputs (wind speed and current power), (d) the RMSE test-set with three inputs (wind speed, direction and current power). }%
\label{fig:box_onehour}
\end{figure}

The same experiment was run for comparing the performance of four models in forecasting the wind turbine power output in the one-hour ahead. The comparative outcomes are presented in Figure \ref{fig:box_onehour}. In the first and second models, the highest accuracy is observed when batch sizes are large; however, the best-found configurations of the hyperparameters in model 3 and 4 are related to the batch size and learning rate of $256$ and $10^{-4}$ respectively.    
\begin{figure}[tbp]
    \centering
     \subfloat[]{
    \includegraphics[width=0.7\columnwidth]{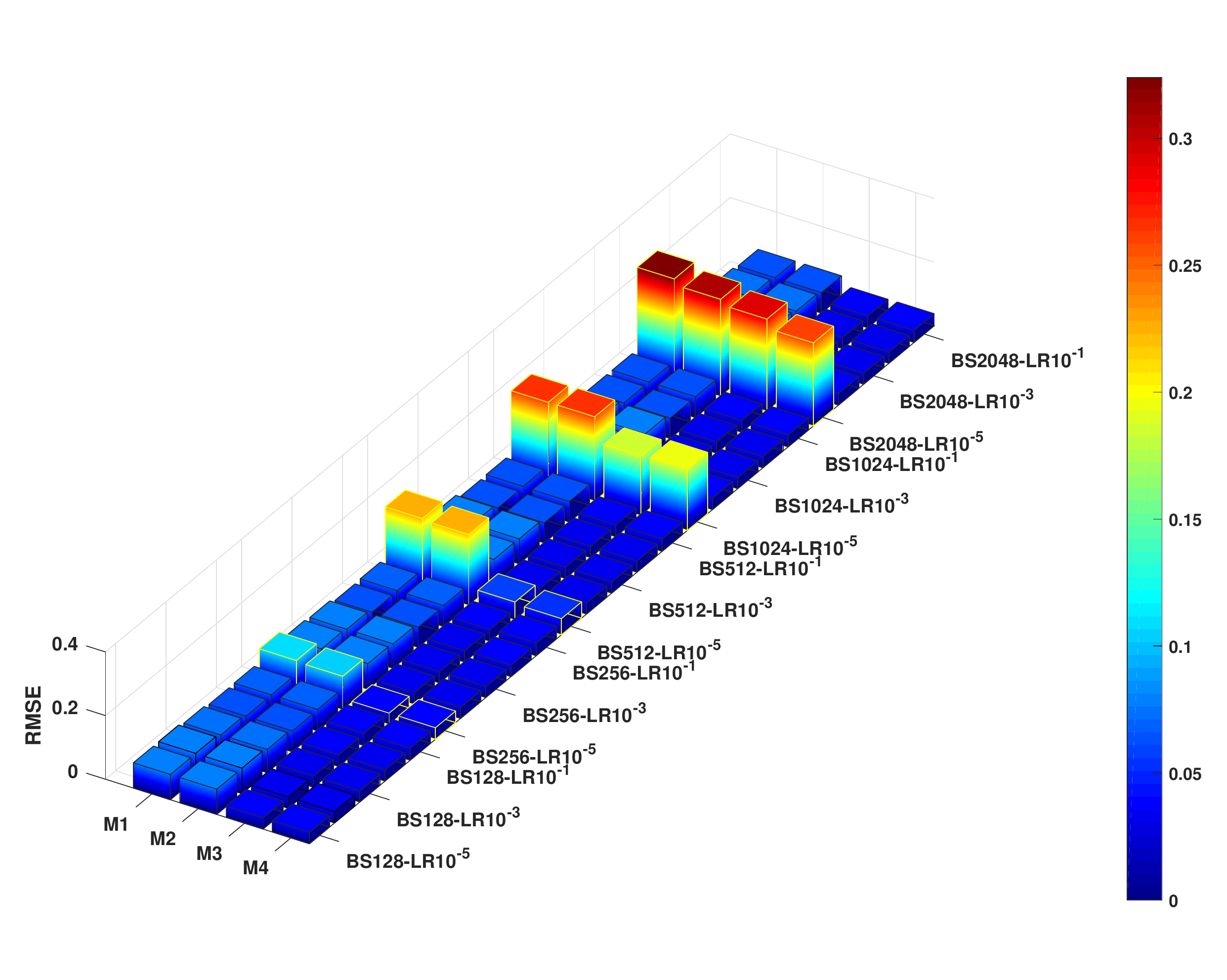}}
    \subfloat[]{
    \includegraphics[width=0.3\columnwidth]{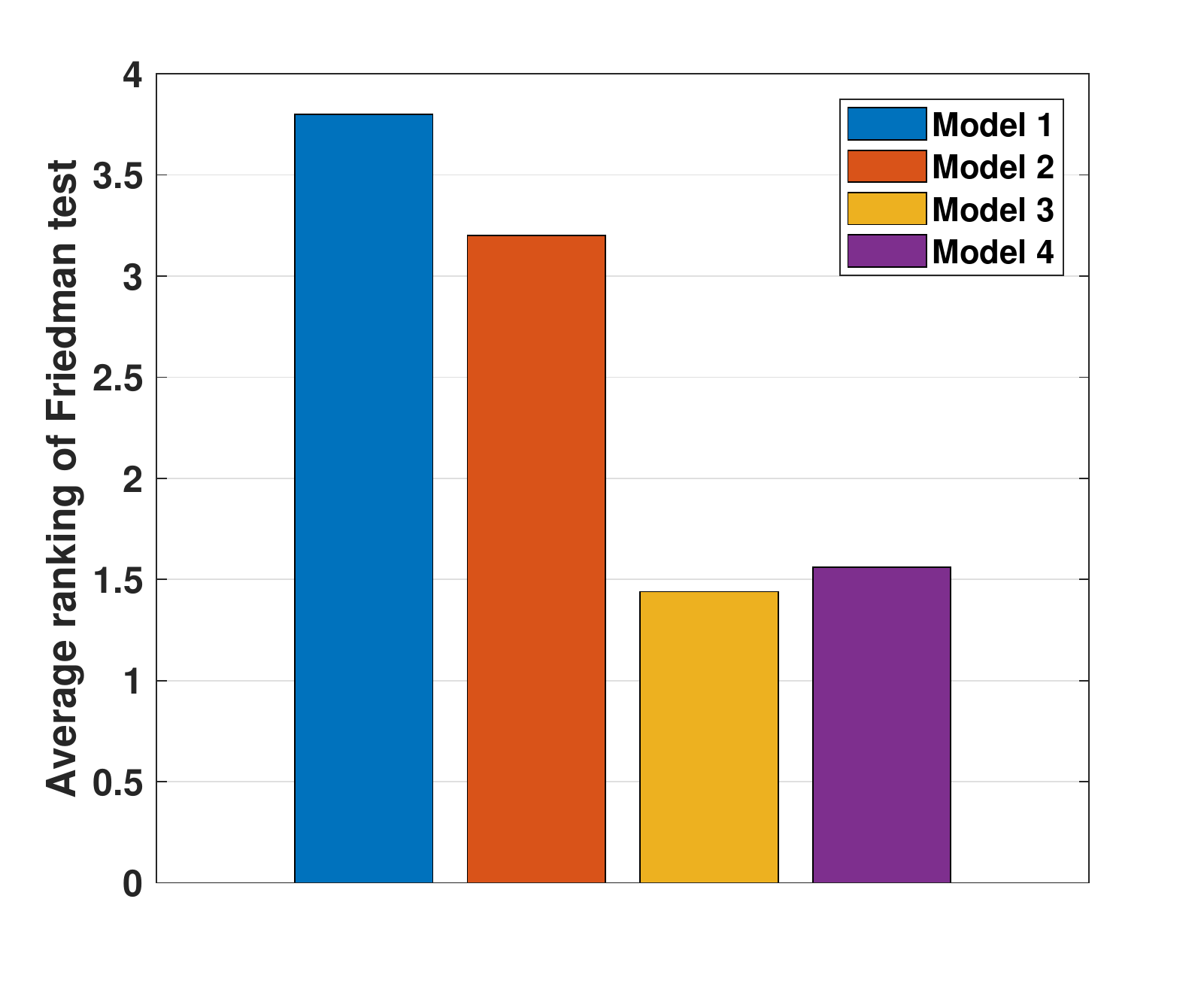}}
     \caption{ A comparison of four proposed forecasting models with 25 different configurations of hyperparameters. (a) A comparison of various LSTM settings based on RMSE. (b) The average ranking of the Friedman test for four applied models.  }
    \label{fig:barplot_friedman}
\end{figure}
We perform a statistical analysis of the achieved forecasting results by the Friedman test (non-parametric and multiple comparisons). The statistical analysis results by implementing the Friedman test are shown in Figure \ref{fig:barplot_friedman}. We have ranked the four forecasting LSTM models corresponding to their mean value. From Figure \ref{fig:barplot_friedman}, Model 3 and 4 obtained the first and second ranking (i.e., the lowest value goes the first rank) compared to all models over the 25 configurations of the hyperparameters under the ten-minute ahead. 
 \begin{figure*}[tbp]
 \centering
 \subfloat[]{
  \includegraphics[width=0.48\textwidth]{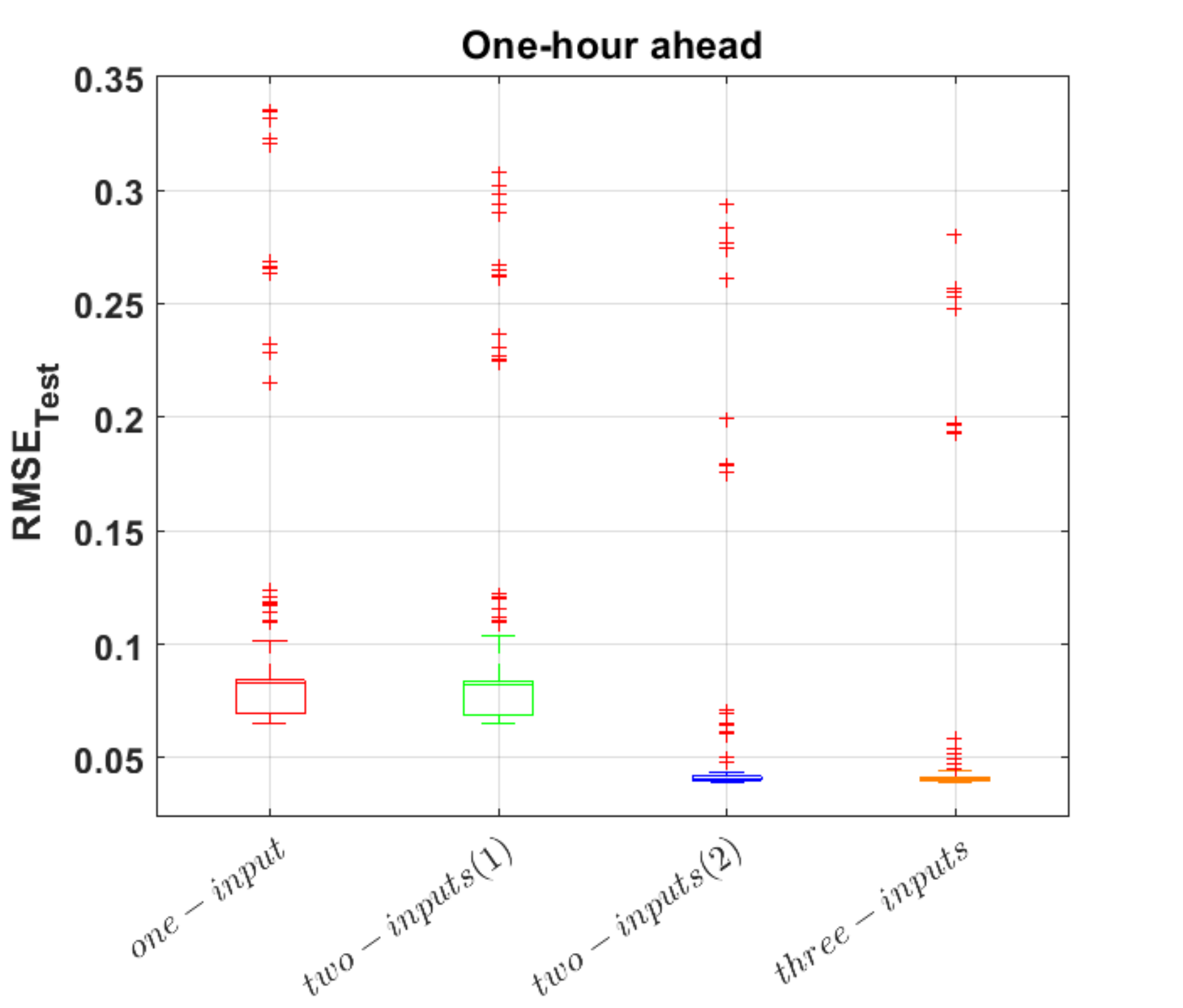}\llap{\raisebox{2.5cm}{\includegraphics[height=3.5cm]{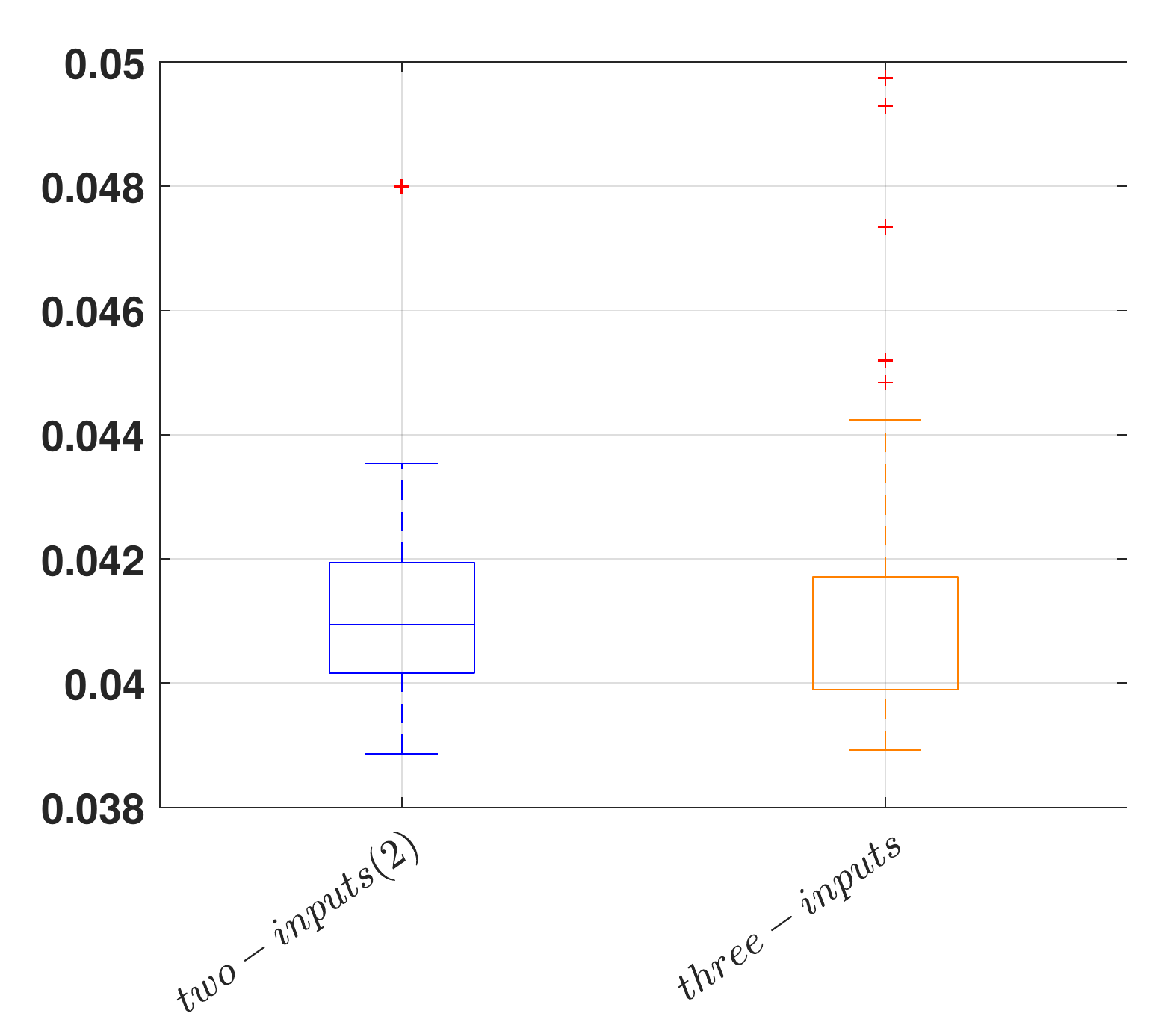}}}}
  \subfloat[]{
  \includegraphics[width=0.48\textwidth]{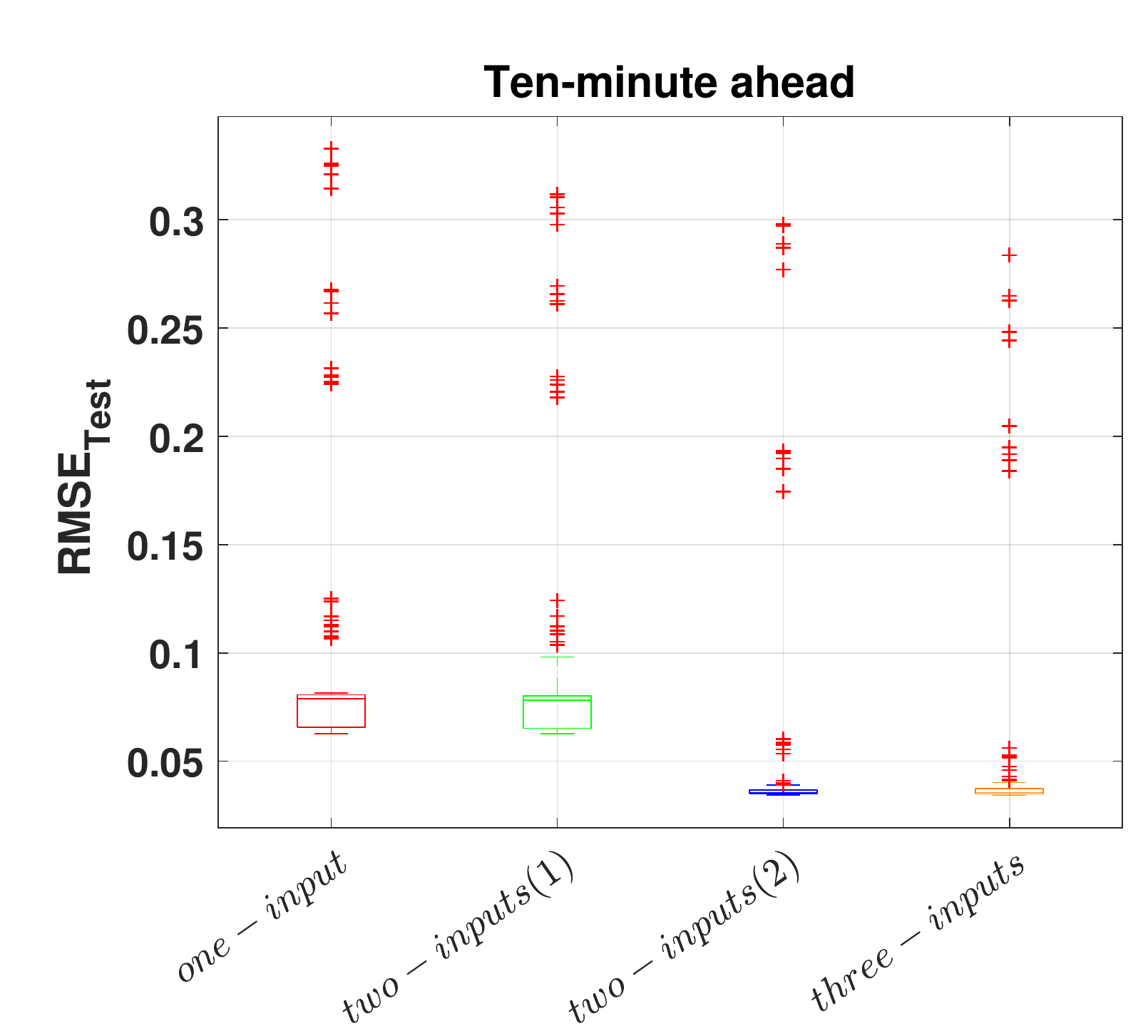}\llap{\raisebox{2.5cm}{\includegraphics[height=3.5cm]{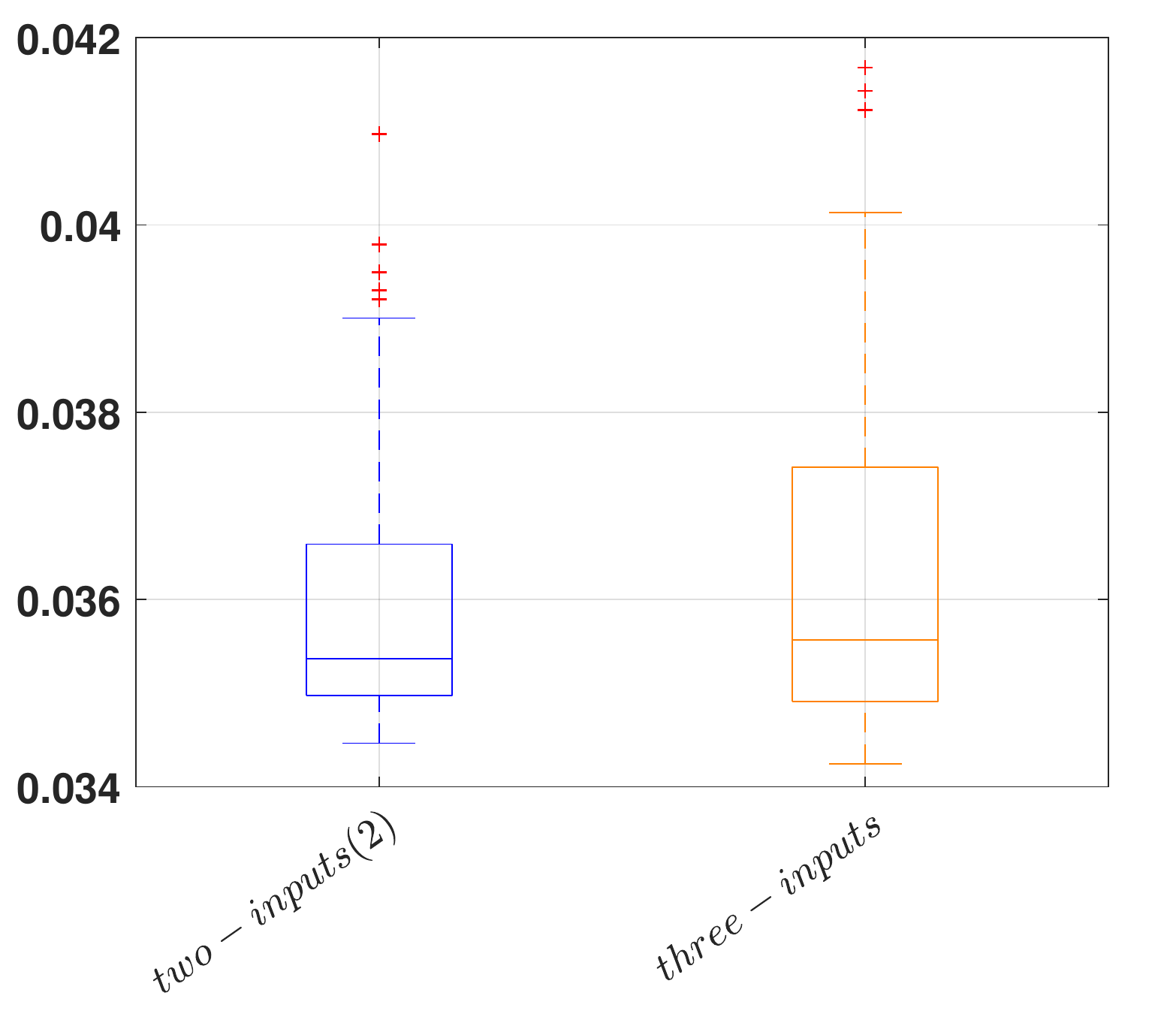}}}}
   \caption{ The total performance comparison of four LSTM forecasting models with ten-minute and one-hour ahead prediction. two-input(1) is the wind speed and direction, two-input(2) mentions the wind speed and current power of wind turbine.}
   \label{fig:boxplot_onehour_ten_all}
 
  \end{figure*}

The overall comparison of four LSTM models on both intervals of ten-minute and one-hour ahead can be seen in Figure \ref{fig:boxplot_onehour_ten_all}. Each box represents the average RMSE forecasting of all 25 configurations per model. These statistical results show that the 4th model with three inputs, including wind speed, wind direction and the current power output of the wind turbine found a configuration with the minimum validation error. However, model 3 outperforms other models on average.   
 \begin{figure*}[tbp]
 \centering
 \subfloat[]{
  \includegraphics[width=.95\textwidth]{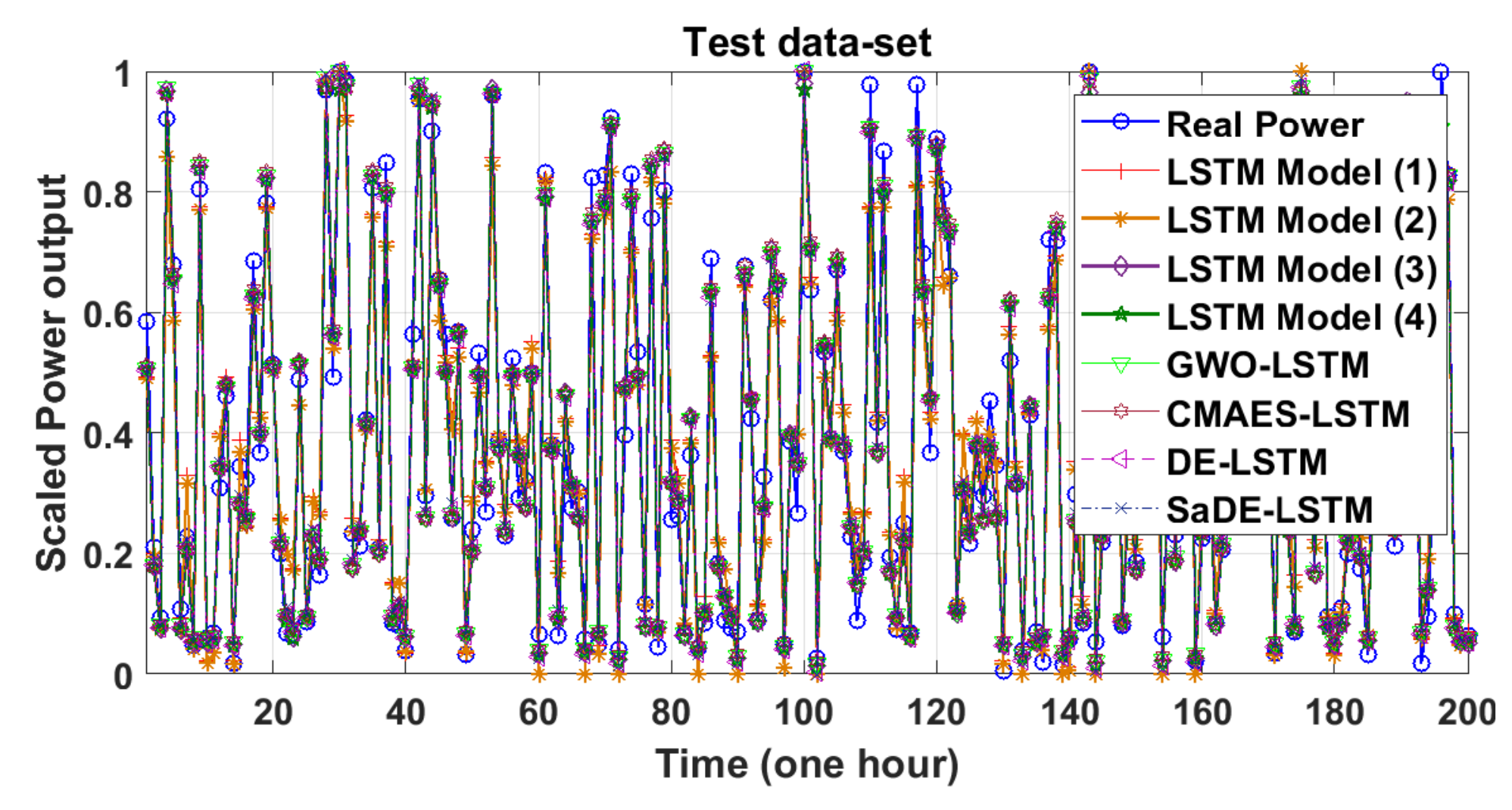}
 }\\
  \subfloat[]{
  \includegraphics[width=0.33\textwidth]{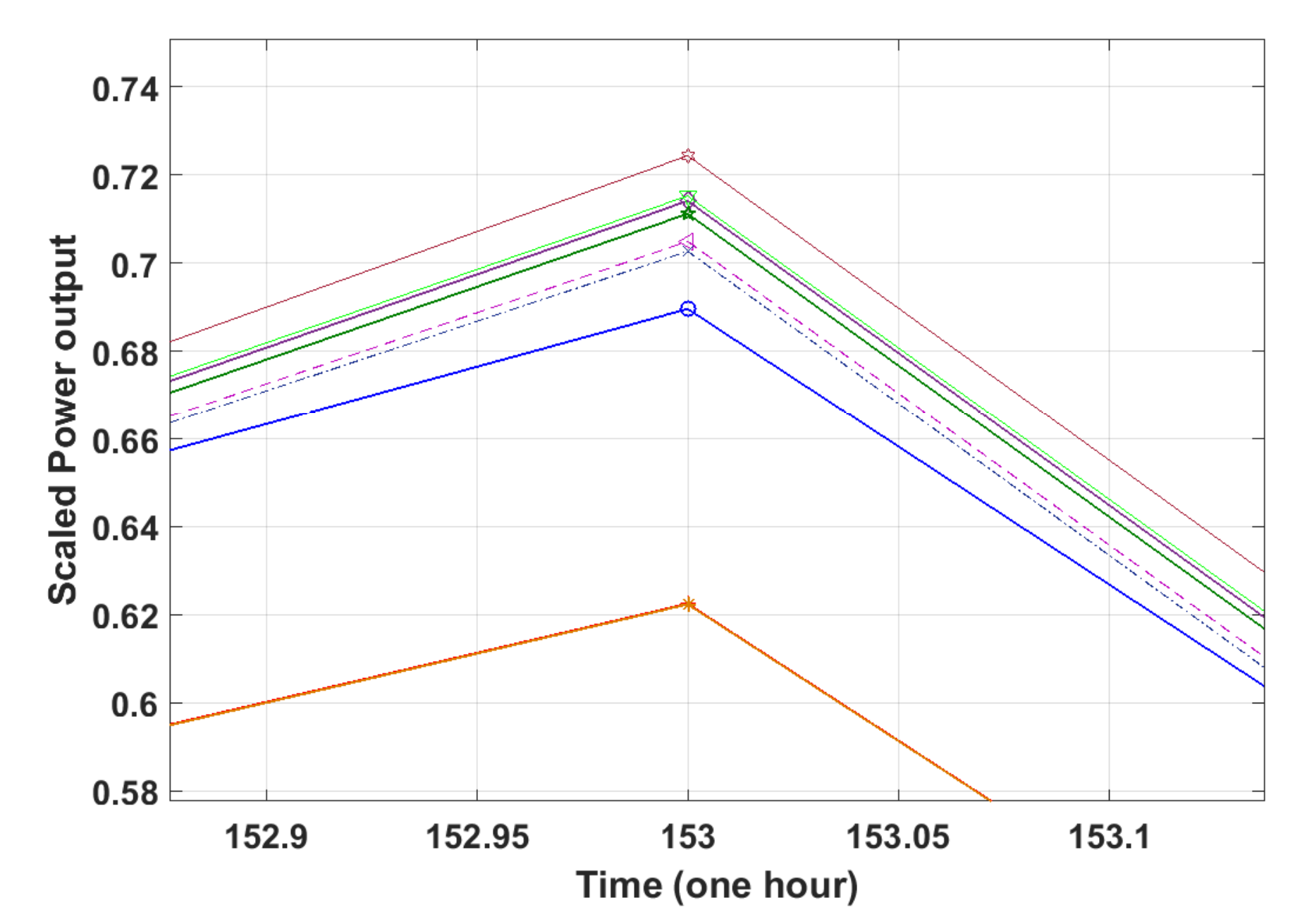}}
  \subfloat[]{
  \includegraphics[width=0.33\textwidth]{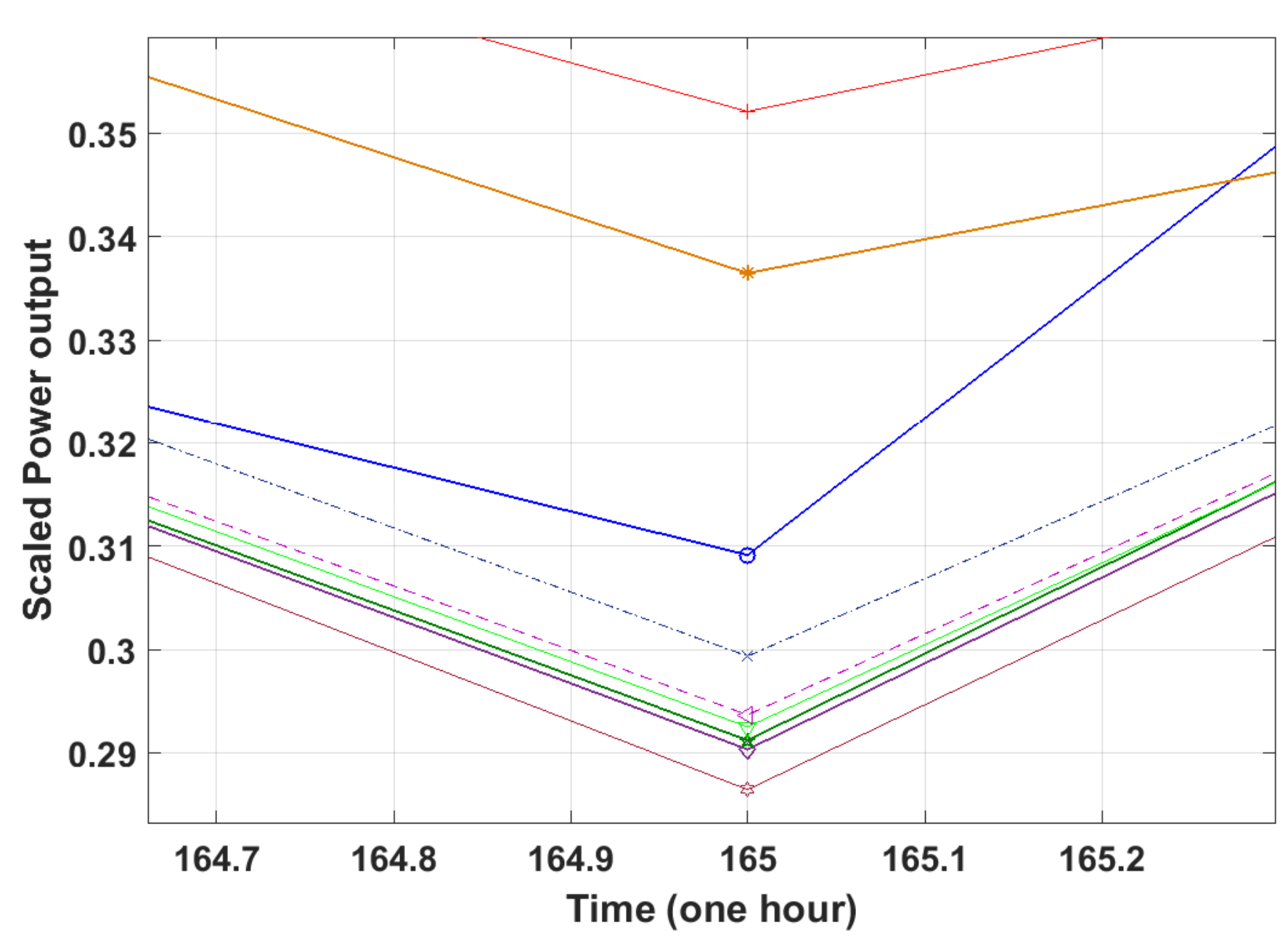}}
  \subfloat[]{
  \includegraphics[width=0.33\textwidth]{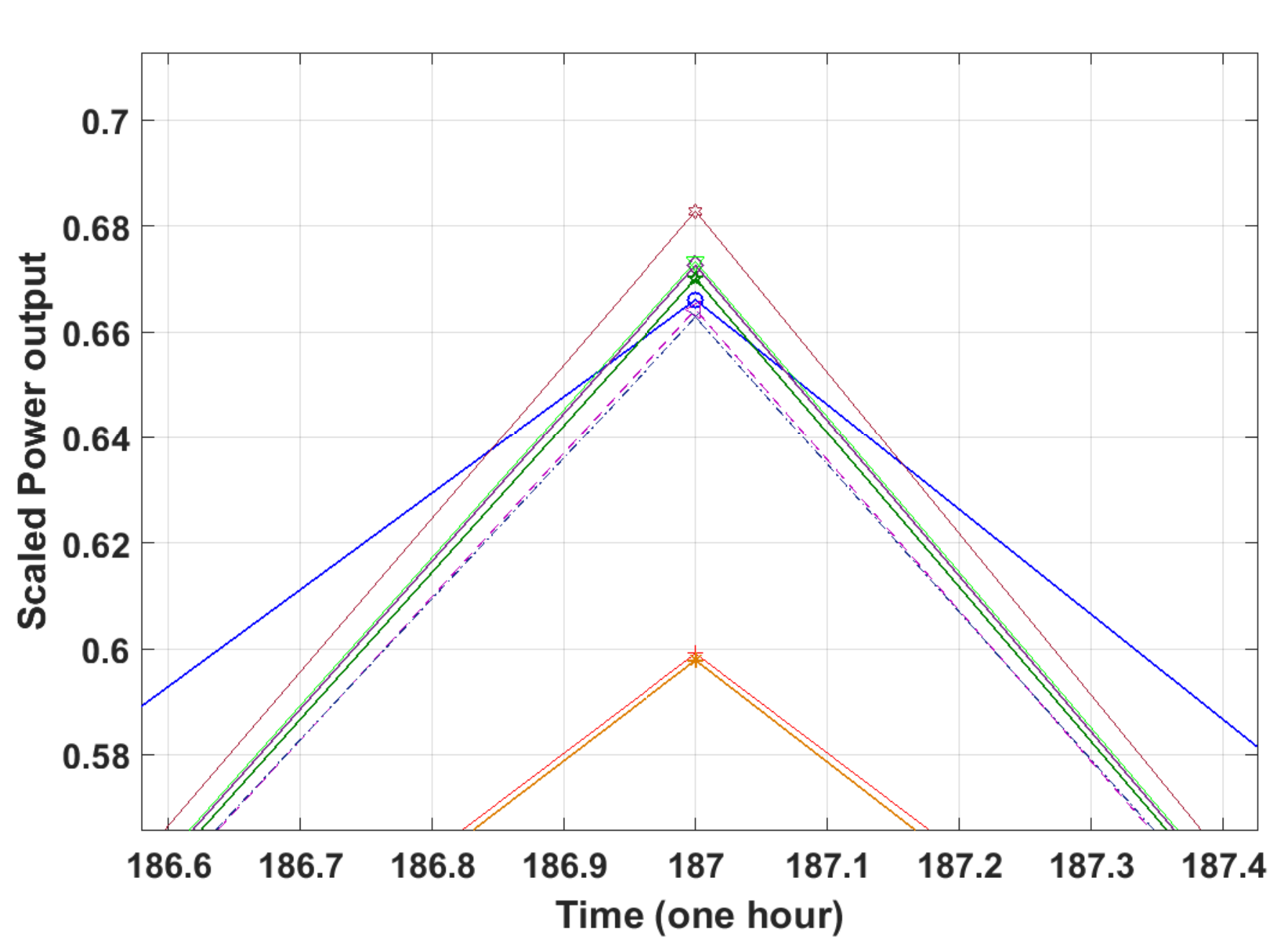}}
 
   \caption{ The best estimated power output values from the proposed hybrid models and the corresponding measured values in SCADA system. The initial values of the weights are kept the same. }
   \label{fig:estimated_power}
 
  \end{figure*}
According to the statistical results, model 3 with two inputs (wind speed and the current produced power) perform better than other models. Therefore, this model is applied to developing hybrid neuro-evolutionary methods. For evaluating the performance of the proposed hybrid model, we compare five different forecasting methods including the best LSTM model which is tuned by the grid search, an adaptive neuro-fuzzy inference system (ANFIS) (its hyperparameters are assigned based on the study in~\cite{pousinho2011hybrid}), and three new hybrid neuro-evolutionary methods (CMAES-LSTM~\cite{neshat2020evolutionary}, DE-LSTM~\cite{peng2018effective} and GWO-LSTM~\cite{mirjalili2015effective,neshat2019adaptive}). 
Table \ref{table:ten-minute} and Table \ref{table:one-hour} summarise the outcomes of the performance indices produced to determine the optimal structure and hyperparameters of the applied forecasters. It is evident that the SaDE-LSTM hybrid model outperforms other hybrid models and provides more accurate forecasting results. Table \ref{table:details_model} reports the best-found configurations of the proposed forecasting model and other compared models.

\begin{table}

\centering
\caption{Performance indices of forecasting wind turbine power output achieved by different models for ten-minutes ahead.}
\scalebox{0.8}{
\begin{tabular}{l|l|l|l|l|l|l|l|l|l}
\hline
\hlineB{4}
 & &   \multicolumn{1}{ c }\textbf{MSE} &  \multicolumn{2}{ c }\textbf{RMSE}    & \multicolumn{2}{ c }\textbf{MAE}& \multicolumn{2}{ c }\textbf{R}          \\  \hline
 
\textbf{Model}&  & \textbf{Train}    & \textbf{Test}     & \textbf{Train}    & \textbf{Test}     & \textbf{Train}    & \textbf{Test}    & \textbf{Train}    & \textbf{Test}     \\  \hline
\textbf{ANFIS}      & Mean & 6.981E-03 & 6.949E-03 & 8.324E-02 & 8.343E-02 & 5.287E-02 & 5.282E-02 & 9.615E-01 & 9.618E-01 \\
           & Min  & 5.657E-03 & 5.725E-03 & 7.566E-02 & 7.521E-02 & 4.795E-02 & 4.809E-02 & 9.593E-01 & 9.584E-01 \\
           & Max  & 7.648E-03 & 7.673E-03 & 8.760E-02 & 8.745E-02 & 5.546E-02 & 5.544E-02 & 9.665E-01 & 9.664E-01 \\
           & Std  & 7.860E-04 & 7.614E-04 & 4.662E-03 & 4.834E-03 & 2.999E-03 & 2.979E-03 & 2.825E-03 & 2.858E-03 \\
            \hlineB{4}  
\textbf{LSTM-grid}  & Mean & 1.427E-03 & 1.407E-03 & 3.778E-02 & 3.751E-02 & 2.834E-02 & 2.819E-02 & 9.826E-01 & 9.829E-01 \\
           & Min  & 1.411E-03 & 1.382E-03 & 3.756E-02 & 3.718E-02 & 2.822E-02 & 2.804E-02 & 9.826E-01 & 9.826E-01 \\
           & Max  & 1.452E-03 & 1.450E-03 & 3.810E-02 & 3.808E-02 & 2.867E-02 & 2.836E-02 & 9.827E-01 & 9.830E-01 \\
           & Std  & 1.520E-04 & 2.552E-04 & 1.233E-02 & 1.598E-02 & 1.908E-03 & 1.216E-04 & 4.968E-05 & 1.606E-04 \\
            \hlineB{4}  
\textbf{CMAES-LSTM} & Mean & 1.236E-03 & 1.193E-03 & 3.515E-02 & 3.454E-02 & 2.643E-02 & 2.615E-02 & 9.926E-01 & 9.928E-01 \\
           & Min  & 1.227E-03 & 1.171E-03 & 3.503E-02 & 3.422E-02 & 2.631E-02 & 2.594E-02 & 9.926E-01 & 9.927E-01 \\
           & Max  & 1.256E-03 & 1.206E-03 & 3.544E-02 & 3.473E-02 & 2.679E-02 & 2.638E-02 & 9.927E-01 & 9.930E-01 \\
           & Std  & 7.280E-06 & 1.161E-05 & 1.033E-04 & 1.682E-04 & 1.206E-04 & 1.393E-04 & 2.597E-05 & 9.001E-05 \\
            \hlineB{4}  
\textbf{DE-LSTM  }  & Mean & 1.241E-03 & 1.163E-03 & 3.522E-02 & 3.411E-02 & 2.648E-02 & 2.583E-02 & 9.926E-01 & 9.930E-01 \\
           & Min  & 1.235E-03 & 1.159E-03 & 3.515E-02 & 3.404E-02 & 2.639E-02 & 2.569E-02 & 9.925E-01 & 9.929E-01 \\
           & Max  & 1.248E-03 & 1.180E-03 & 3.533E-02 & 3.436E-02 & 2.667E-02 & 2.609E-02 & 9.926E-01 & 9.932E-01 \\
           & Std  & 4.002E-06 & 5.727E-06 & 5.680E-05 & 8.376E-05 & 7.153E-05 & 1.087E-04 & 2.693E-05 & 6.738E-05 \\
            \hlineB{4}  
\textbf{GWO-LSTM}   & Mean & 1.586E-03 & 1.495E-03 & 3.983E-02 & 3.866E-02 & 2.944E-02 & 2.830E-02 & 9.912E-01 & 9.911E-01 \\
           & Min  & 1.404E-03 & 1.403E-03 & 3.747E-02 & 3.745E-02 & 2.932E-02 & 2.815E-02 & 9.909E-01 & 9.904E-01 \\
           & Max  & 1.631E-03 & 1.549E-03 & 4.038E-02 & 3.936E-02 & 2.977E-02 & 2.847E-02 & 9.913E-01 & 9.914E-01 \\
           & Std  & 2.198E-03 & 1.154E-03 & 4.688E-02 & 3.397E-02 & 1.908E-03 & 1.216E-04 & 1.309E-04 & 3.870E-04 \\
            \hlineB{4}  
\textbf{SaDE-LSTM } & Mean & 1.167E-03 & 1.133E-03 & 3.414E-02 & 3.365E-02 & 2.542E-02 & 2.504E-02 & 9.931E-01 & 9.935E-01 \\
           & Min  & 1.006E-03 & 1.016E-03 & 3.171E-02 & 3.187E-02 & 2.320E-02 & 2.299E-02 & 9.928E-01 & 9.930E-01 \\
           & Max  & 1.244E-03 & 1.210E-03 & 3.528E-02 & 3.478E-02 & 2.650E-02 & 2.649E-02 & 9.936E-01 & 9.941E-01 \\
           & Std  & 8.014E-05 & 6.429E-05 & 1.192E-03 & 9.620E-04 & 8.965E-04 & 8.936E-04 & 2.257E-04 & 4.043E-04\\   
\end{tabular}
}
\label{table:ten-minute}
\end{table}

\begin{table}

\centering
\caption{Performance indices of forecasting wind turbine power output achieved by different models for one-hour ahead.}
\scalebox{0.8}{
\begin{tabular}{l|l|l|l|l|l|l|l|l|l}
\hline
\hlineB{4}
&  &  \multicolumn{1}{ c }\textbf{MSE} &  \multicolumn{2}{ c }\textbf{RMSE}    & \multicolumn{2}{ c }\textbf{MAE}& \multicolumn{2}{ c }\textbf{R}          \\  \hline
 
\textbf{Model}&  & \textbf{Train}    & \textbf{Test}     & \textbf{Train}    & \textbf{Test}     & \textbf{Train}    & \textbf{Test}    & \textbf{Train}    & \textbf{Test}     \\  \hline
\textbf{ANFIS}      & Mean & 6.875E-03 & 6.789E-03 & 8.284E-02 & 8.229E-02 & 5.319E-02 & 5.289E-02 & 9.608E-01 & 9.610E-01 \\
           & Min  & 6.208E-03 & 6.013E-03 & 7.879E-02 & 7.754E-02 & 5.071E-02 & 4.986E-02 & 9.585E-01 & 9.575E-01 \\
           & Max  & 7.617E-03 & 7.905E-03 & 8.728E-02 & 8.891E-02 & 5.620E-02 & 5.711E-02 & 9.634E-01 & 9.636E-01 \\
           & Std  & 6.226E-04 & 7.344E-04 & 3.758E-03 & 4.421E-03 & 2.348E-03 & 2.851E-03 & 2.318E-03 & 2.314E-03 \\
           \hlineB{4}  
\textbf{LSTM-grid}  & Mean & 1.906E-03 & 1.909E-03 & 4.366E-02 & 2.089E-01 & 3.046E-02 & 3.053E-02 & 9.903E-01 & 9.905E-01 \\
           & Min  & 1.880E-03 & 1.827E-03 & 4.336E-02 & 2.082E-01 & 3.036E-02 & 3.024E-02 & 9.902E-01 & 9.899E-01 \\
           & Max  & 1.923E-03 & 2.004E-03 & 4.386E-02 & 2.094E-01 & 3.055E-02 & 3.083E-02 & 9.905E-01 & 9.909E-01 \\
           & Std  & 1.968E-05 & 8.082E-05 & 2.458E-04 & 1.004E-03 & 6.857E-05 & 2.415E-04 & 1.238E-04 & 4.684E-04 \\
           \hlineB{4}  
\textbf{CMAES-LSTM} & Mean & 1.634E-03 & 1.547E-03 & 4.042E-02 & 3.933E-02 & 3.061E-02 & 3.033E-02 & 9.902E-01 & 9.907E-01 \\
           & Min  & 1.608E-03 & 1.520E-03 & 4.010E-02 & 3.898E-02 & 3.041E-02 & 3.000E-02 & 9.901E-01 & 9.905E-01 \\
           & Max  & 1.652E-03 & 1.589E-03 & 4.065E-02 & 3.987E-02 & 3.084E-02 & 3.066E-02 & 9.903E-01 & 9.910E-01 \\
           & Std  & 1.208E-05 & 2.141E-05 & 1.495E-04 & 2.717E-04 & 1.180E-04 & 1.854E-04 & 6.810E-05 & 1.496E-04 \\
           \hlineB{4}  
\textbf{DE-LSTM }   & Mean & 1.645E-03 & 1.483E-03 & 4.056E-02 & 3.851E-02 & 3.064E-02 & 3.005E-02 & 9.901E-01 & 9.911E-01 \\
           & Min  & 1.636E-03 & 1.467E-03 & 4.045E-02 & 3.830E-02 & 3.050E-02 & 2.963E-02 & 9.901E-01 & 9.909E-01 \\
           & Max  & 1.655E-03 & 1.519E-03 & 4.068E-02 & 3.897E-02 & 3.085E-02 & 3.026E-02 & 9.902E-01 & 9.913E-01 \\
           & Std  & 6.622E-06 & 1.504E-05 & 8.163E-05 & 1.948E-04 & 1.043E-04 & 1.679E-04 & 3.803E-05 & 1.328E-04 \\
           \hlineB{4}  
\textbf{GWO-LSTM}   & Mean & 1.981E-03 & 1.999E-03 & 4.451E-02 & 4.471E-02 & 3.050E-02 & 3.053E-02 & 9.887E-01 & 9.887E-01 \\
           & Min  & 1.870E-03 & 1.887E-03 & 4.324E-02 & 4.344E-02 & 3.037E-02 & 3.034E-02 & 9.882E-01 & 9.882E-01 \\
           & Max  & 1.202E-03 & 1.904E-03 & 3.467E-02 & 4.364E-02 & 3.055E-02 & 3.097E-02 & 9.890E-01 & 9.894E-01 \\
           & Std  & 2.675E-05 & 2.082E-05 & 5.172E-03 & 4.563E-03 & 6.857E-05 & 4.415E-04 & 2.948E-04 & 4.635E-04 \\
           \hlineB{4}  
\textbf{SaDE-LSTM } & Mean & 1.431E-03 & 1.413E-03 & 3.779E-02 & 3.755E-02 & 2.884E-02 & 2.833E-02 & 9.919E-01 & 9.921E-01 \\
           & Min  & 1.242E-03 & 1.237E-03 & 3.525E-02 & 3.517E-02 & 2.761E-02 & 2.757E-02 & 9.911E-01 & 9.912E-01 \\
           & Max  & 1.636E-03 & 1.632E-03 & 4.045E-02 & 4.040E-02 & 2.968E-02 & 2.896E-02 & 9.932E-01 & 9.931E-01 \\
           & Std  & 1.457E-04 & 1.394E-04 & 1.923E-03 & 1.850E-03 & 6.034E-04 & 4.214E-04 & 6.733E-04 & 6.051E-04
\end{tabular}
}
\label{table:one-hour}
\end{table}

In addition, The actual and forecasting values generated by the hybrid models and the LSTM networks for the one-hour ahead forecasting are shown in Figure \ref{fig:estimated_power}. In the zoomed versions of Figure \ref{fig:estimated_power}(b, c and d), it can be seen that the SaDE-LSTM estimates the power output with considerable accuracy compared with other models. 

 \section{Conclusions} \label{sec:Conclusions}
Due to multiple systems and meteorological factors, wind power time series data exhibit chaotic behaviours which are hard to predict. In this paper, a combination of autoencoder and clustering was adopted to reduce the stochastic noise inherent in raw data series. Subsequently, a neuro-evolutionary approach (SaDE-LSTM)  consisting of the self-adaptive differential evolution (SaDE) and LSTM network was used for modelling the wind behaviour. We then conducted extensive experiments and compared our proposed approach with five alternative hybrid models. As our experiments suggest, the proposed SaDE-LSTM model outperforms its counterparts in terms of four performance criteria, in both ten-minute and one-hour intervals.

In the future, more power curve datasets derived from different kinds of wind turbines from different regions will be evaluated to enhance our model further. Ultimately, another investigation of this study is to employ various outlier detection methods and optimization approaches to improve the forecasting results.


\bibliographystyle{unsrt}  
\bibliography{sample-bibliography}
\end{document}